\documentclass[runningheads]{llncs}
\usepackage[T1]{fontenc}
\usepackage{graphicx}
\usepackage{xcolor}
\usepackage[colorlinks]{hyperref}
\usepackage{graphicx}
\usepackage{amsmath}
\usepackage{amssymb}
\usepackage{booktabs}
\usepackage{caption}
\usepackage{subcaption}
\usepackage{multirow}
\usepackage{microtype}
\usepackage{adjustbox}

\def\mD{\mathcal{D}}

\def\mL{\mathcal{L}}

\def\1n{\mathbf{1}_n}
\def\0{\mathbf{0}}
\def\1{\mathbf{1}}

\def\x{{\bf x}}
\def\y{{\bf y}}
\def\z{{\bf z}}

\definecolor{pink}{rgb}{0.9,0.5,0.5}
\definecolor{purple}{rgb}{0.5, 0.4, 0.8}   
\definecolor{gray}{rgb}{0.3, 0.3, 0.3}
\definecolor{mygreen}{rgb}{0.2, 0.6, 0.2}

\definecolor{greena}{rgb}{0.4, 0.5, 0.1}

\definecolor{bluea}{rgb}{0, 0.4, 0.6}

\definecolor{reda}{rgb}{0.6, 0.2, 0.1}

\newcommand{\cm}[1]{}

\newcommand{\myheading}[1]{\vspace{1ex}\noindent \textbf{#1}}

\newif\ifshowsolution
\showsolutiontrue

\ifshowsolution

\else

\fi

\newcommand{\Sref}[1]{Sec.~\ref{#1}}
\newcommand{\Eref}[1]{Eq.~(\ref{#1})}
\newcommand{\Fref}[1]{Fig.~\ref{#1}}
\newcommand{\Tref}[1]{Table~\ref{#1}}

\def\fsim{\psi}
\def\x{\mathbf{x}}
\def\y{\mathbf{y}}
\def\z{\mathbf{z}}

\begin{document}
\title{Self-Supervised Learning with Multi-View Rendering for 3D Point Cloud Analysis}
\titlerunning{SSL with Multi-View Rendering for 3D Point Cloud Analysis}
%
\author{Bach Tran\inst{1} \and
Binh-Son Hua\inst{1} \and
Anh Tuan Tran\inst{1} \and
Minh Hoai\inst{1,2}}
\authorrunning{B. Tran et al.}
%
\institute{VinAI Research, Hanoi, Vietnam \and
Stony Brook University, New York, 11794, USA
\email{\{v.bachtx12,v.sonhb,v.anhtt152,v.hoainm\}@vinai.io}}
\maketitle              
\begin{abstract}
Recently, great progress has been made in 3D deep learning with the emergence of deep neural networks specifically designed for 3D point clouds. These networks are often trained from scratch or from pre-trained models learned purely from point cloud data. Inspired by the success of deep learning in the image domain, we devise a novel pre-training technique for better model initialization by utilizing the multi-view rendering of the 3D data. Our pre-training is self-supervised by a local pixel/point level correspondence loss computed from perspective projection and a global image/point cloud level loss based on knowledge distillation, thus effectively improving upon popular point cloud networks, including PointNet, DGCNN and SR-UNet. 
These improved models outperform existing state-of-the-art methods on various datasets and downstream tasks. We also analyze the benefits of synthetic and real data for pre-training, and observe that pre-training on synthetic data is also useful for high-level downstream tasks. 
Code and pre-trained models are available at \href{https://github.com/VinAIResearch/selfsup_pcd.git}{\url{https://github.com/VinAIResearch/selfsup_pcd.git}}.

\end{abstract}
\section{Introduction}

Pixels and points are basic elements in computer vision for visual recognition. 
In the past decade, image collections have been successfully used to train neural networks for common visual recognition tasks, including object classification and semantic segmentation. 
Concurrently, advances in depth-sensing technologies, including RGB-D and LiDAR sensors, have enabled the acquisition of large-scale 3D data, facilitating the rapid development of visual recognition methods in 3D, notably neural networks for point cloud analysis in the last few years.
Unlike images, annotation for point clouds are generally more expensive to acquire due to the laborious process of scene scanning, reconstruction, and annotation, and thus point cloud neural networks are often trained with datasets that are much smaller than image datasets. 

A potential direction to improve the robustness for point cloud neural networks is self-supervised learning. 
By letting the point cloud network perform some pre-text tasks before supervised learning, a process commonly known as pre-training, the network can perform more effectively than that trained from scratch. 
With self-supervised learning, the pre-text tasks are designed so that the pre-training does not use additional labels. 
In 3D deep learning, some initial effort has been spent on investigating this direction~\cite{sauder2019self,wang2021unsupervised,xie2020pointcontrast}. 
However, most previous works solely considered self-supervised learning in the 3D domain; only a few works exploited images to support the learning of point cloud neural networks. 
In an early work, Pham et al.~\cite{pham2020lcd} trained autoencoders on both images and point clouds and applied constraints on the latent space of both domains, allowing feature transfers between 2D and 3D.
Inspired by the recently growing literature on network pre-training, we explore how to use images to more effectively (self-)supervise point cloud neural networks.

Particularly, in this paper, we propose a method that utilizes multi-view rendering to generate pixel/point and image/point cloud pairs for self-supervising a point cloud neural network. 
We train two neural networks, one for image and one for point cloud, respectively, and direct both networks to agree upon their latent features in the 2D and 3D domains.
To achieve this, we use the pixel and point correspondences to formulate a local loss function that encourages features of the correspondences to be similar. 
This is well-motivated by projective geometry in 3D computer vision that defines the coordinate mapping between the image and 3D space.  
To further regularize the self-supervision, we utilize knowledge distillation to formulate a global loss that encourages the feature distribution between images and point clouds to be similar as well. Our method works even when there is big domain gap between the pre-train data and test data, e.g., between synthetic and real data. 


In summary, we make three technical contributions in this paper: (1) a pre-training technique built upon multi-view rendering that advocates the use of multi-view image features to self-supervise the point cloud neural network; (2) a local loss function that exploits pixel-point correspondence in the pre-training; (3) a global loss function that performs knowledge distillation from the images to the 3D point clouds.

\section{Related work}
\myheading{3D deep learning:} 
Building a neural network to analyze 3D data is a non-trivial task. Early attempts involve extending neural networks for images to work with volumes~\cite{maturana2015voxnet}, and projecting 3D data to 2D views that can be used with traditional neural networks~\cite{su2015multi}. 
Recent methods in deep learning with point clouds take a different approach by letting a network input point clouds directly. 
Two major directions can be taken to implement such idea: learning per-point features by pointwise MLP~\cite{Qi_2017_CVPR,qi2017pointnet++}, and learning point features by examining points in a local neighborhood by custom convolutions tailored for point sets~\cite{liu2019relation,hua2018pointwise,li2018pointcnn} and by graph neural networks~\cite{wang2019dynamic,simonovsky2017dynamic,bruna2013spectral}. 
Notable methods in such directions include PointNet~\cite{Qi_2017_CVPR} and DGCNN~\cite{wang2019dynamic}. 
An inherent limitation of PointNet-based approaches is that they can only process a few thousands of points, limiting the ability to handle large-scale point clouds, where a sliding window is often used as a workaround~\cite{Qi_2017_CVPR}. 
More recent developments include the use of sparse tensor and sparse convolution~\cite{choy20194d,graham20183d,graham2017submanifold} on large-scale point clouds for semantic segmentation and 3D object detection. We refer readers to \cite{guo2020deep} for a survey of deep learning methods for point clouds.


\myheading{Self-supervised learning:} 
Unsupervised pre-training is a useful technique in deep learning with proven success in natural language processing \cite{devlin2018bert} and representation learning \cite{chen2020simple,grill2020bootstrap,he2020momentum,misra2020self,zhang2019unsupervised,sharma2020self,thabet2020self,hassani2019unsupervised}. 
For pre-training, one can use generative modeling techniques based on GANs \cite{wu2016learning,achlioptas2018learning} and autoencoders \cite{hassani2019unsupervised,wang2021unsupervised,han2019multi,yu2022point}, or other self-supervised learning techniques~\cite{xie2020pointcontrast,sauder2019self,chen2021shape,poursaeed2020self,hou2021exploring,yamada2022point,alliegro2021joint,achituve2021self}. 
Pre-training is relevant to knowledge distillation \cite{hinton2015distilling}, a class of techniques for transferring features learned from a large network (teacher network) to a small network (student network). Here we assume that the pre-text task is rather general and can be very different to the downstream tasks, and so a subset of the layers in the pre-trained can be transferred depending on the downstream task. 


Self-supervised learning techniques for pre-training 3D point cloud networks have been recently explored from multiple perspectives. 
Early works use a pre-text task for self-supervised learning.
The pre-text task can be solving a jigsaw puzzle~\cite{sauder2019self}, where a point cloud is subdivided into randomly arranged voxels, and the task is to predict for each point the voxel ID the point belongs to.
Another pre-text task is point cloud completion~\cite{wang2021unsupervised} (OcCo), where a mechanism similar to mask-based pre-training in natural language processing is utilized.
As an extension of autoencoder on 3D point clouds, Eckart et al.~\cite{eckart2021self} apply soft segmentation on point clouds and enforces these partitions to comply a latent parametric model in an encoder-decoder network paradigm. 
Recent contrastive learning is also shown to be effective for pre-training 3D point clouds~\cite{xie2020pointcontrast,zhang2021self}.
PointContrast~\cite{xie2020pointcontrast} create positive pairs and negative pairs for contrastive learning by the correspondences between two camera views of a point cloud. 
DepthContrast~\cite{zhang2021self} learn the representation with multiple 3D data formats including points and voxels. 

Self-supervised learning with other 3D data modalities~\cite{gupta2016cross,liu20213d,hafner2018cross,Hou_2021_ICCV,liu2021learning,afham2022crosspoint,pham2020lcd} has also been explored.
Jing et al.~\cite{jing2020self,jing2021self} (CM) use 3D data with multi-modality and build cross-modal and cross-view invariant constraints, maximizing cross-modal agreement of the features of point cloud, mesh, and images, and maximizing cross-view agreement with the image features. 
Hou et al.~\cite{Hou_2021_ICCV} use contrastive learning on multi-view images constraints and image-geometry constraint. However, they only focus on 2D downstream tasks.
Huang et al.~\cite{huang2021spatio} (STRL) proposed self-supervised learning for a sequence of point clouds which utilizes spatio-temporal cues.
Pham et al.~\cite{pham2020lcd} (LCD) leverages a 2D-3D correspondence dataset and a triplet loss to transfer features from 2D to 3D only on \emph{small cropped regions} of images and 3D point clouds. 
Compared with LCD~\cite{pham2020lcd}, our method is largely different as we self-supervise 3D point features via multi-view projection in the \emph{entire} image space. 
LCD~\cite{pham2020lcd} is suitable for image matching and point cloud registration tasks, while our method is suitable for point cloud analysis tasks such as classification and segmentation. 

There are a few concurrent works~\cite{afham2022crosspoint,li2022simipu}.
In~\cite{afham2022crosspoint}, the authors considered RGB rendering of the object surfaces but required the mesh textures in addition to the geometry for rendering. Our rendering is more practical in that we consider different rendering techniques and only require colorless point clouds. 
In~\cite{li2022simipu}, the authors focus on data from autonomous driving datasets including KITTI and nuScenes. Our method focuses on object datasets. 


\section{Self-supervised learning for 3D point clouds}

\begin{figure*}[t]
\begin{center}
\includegraphics[width=\textwidth]{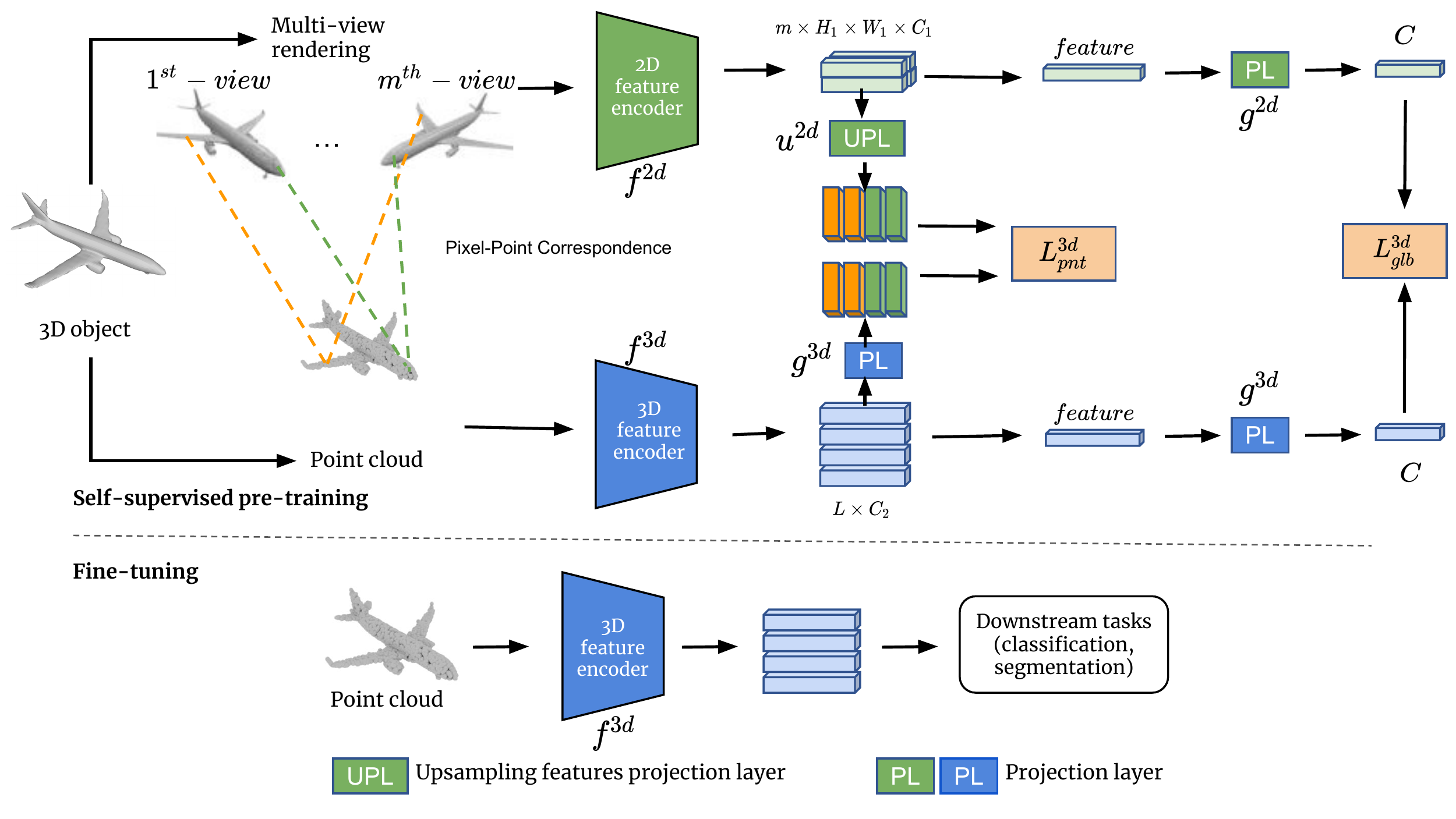}
\caption{Overview of our proposed method. The main proposal is pre-training steps that exploit multi-modal data, including multi-view images and point clouds, to learn a 3D feature encoder for effective point cloud representation. This model is then fine-tuned for different downstream tasks.}
\label{fig:model}
\end{center}
\vskip -0.3in
\end{figure*}

In this section, we describe the proposed self-supervised learning with multi-view rendering for point clouds. 
Our goal is to leverage multi-modal data of 3D objects, in which each object is associated with a 3D point cloud and multiple 2D images from various view points to pre-train the point cloud network. 
We propose to use multi-view rendering to generate images for input 3D objects that pair with the point clouds for pre-training. 
Using rendered images to pre-train point cloud networks implies an advantage that different 3D data representations, including triangle mesh and point cloud, can all be converted into a unified 2D representation to pre-train the networks. 
To ease this pre-training process, we do not require annotation for the 3D objects and rely on self-supervised learning techniques for the pre-training. 

Our method consists of two steps. First, we learn feature representation for 2D images with self-supervised learning by ensuring the similarity between the representation features of two transformed images of the same original image. Second, we use the feature representation of 2D images to learn the 3D feature representation for 3D point clouds. 
We illustrate the overview of our method in \Fref{fig:model}, and we will describe the two steps in details in the next two subsections.

Let $\mD$ denote the pre-training data, $\mD = \{P_i, \{\y_{ij}, M_{ij}\}^{m}_{j=1}\}_{i=1}^{n}$, where $n$ is the number of objects in the dataset and $m$ the number of 2D views for each object. $P_i$ is the 3D point cloud of the $i^{th}$ object, $\y_{ij}$ is the projected image of the $i^{th}$ object to the $j^{th}$ view using the projection matrix $M_{ij}$.

\subsection{Learning feature representation for 2D images}\label{sec:step1}

Let us start with learning the discriminative feature representation for multi-view images. 
In this step, we simply treat all object views $\{\{\y_{ij}\}^{m}_{j=1}\}_{i=1}^{n}$ as items of a set.
Following SimCLR~\cite{chen2020simple}, we randomly sample a batch of $k$ images from this image set in each training iteration. For each image in the batch, we randomly sample two augmentation operators (crop, color distortion, and Gaussian blur) to generate a pair of correlated images based on the original image.
Given an image $\x_i$ in the batch, let $\x_i'$ and $\x_i''$ be its augmented images, respectively. 
Our objective is to learn a feature extraction network $f^{2d}$ so that the resulting feature vectors for different augmentations of the same image $\x_i'$ and $\x_i''$ are similar, while both $\x_i'$ and $\x_i''$ should be different from the feature vectors of other image augmentations $\x_j'$ and $\x_j''$ for $j \neq i$. We therefore define the loss for image $\x_i$ as:
\begin{align}
     \mL^{2d}(i) = &-\log\left(\frac{\fsim(\x_i', \x_i'') }{\fsim(\x_i', \x_i'') + \sum_{j\neq i} (\fsim(\x_i', \x_j') +  \fsim(\x_i', \x_j''))}\right) \\
     &-\log\left(\frac{\fsim(\x_i'', \x_i') }{\fsim(\x_i'', \x_i') + \sum_{j\neq i} (\fsim(\x_i'', \x_j') +  \fsim(\x_i'', \x_j''))}\right). \nonumber 
\end{align}
Here, $\fsim(\x_i, \x_j)$ is the function that measures the similarity between two images, and we use the exponential cosine similarity of the two feature vectors, i.e., 
\begin{align}
    \fsim(\x_i, \x_j) = \exp\Big(\cos\left(g^{2d}(f^{2d}(\x_i)), g^{2d}(f^{2d}(\x_j)) \right) /\tau \Big),
\label{eq:sim}
\end{align} where $\tau$ is the temperature hyper-parameter, and $g^{2d}$ is the projection layer (nonlinear projection layer). 

The loss function for a batch of $k$ images is: $\mL^{2d} = \frac{1}{k}\sum_{i=1}^{k}\mL^{2d}(i)$. 
In each optimization iteration, we calculate the gradient of this loss to optimize for the parameters of the feature extractor network $f^{2d}$, which is a fully convolutional neural network. The input to the network is an RGB image of dimensions $H{\times}W{\times}3$ and the output is a 3D tensor of size $H_1{\times}W_1{\times}C_1$, where $C_1$ is the number of output channels and $H_1 = H/2^s, W_1 = W/2^s$, with $s$ being the number of down-sampling layers in the network. The output tensor can be vectorized to form a global representation vector for the entire image. This output tensor can also be up-sampled to yield a feature map having the same width and height as those of the input image; in this case,  there is a corresponding $C_1$-dim feature vector for each pixel of the input image.





\subsection{Knowledge transfer from 2D to 3D}\label{sec:step2}

Once the feature extraction function $f^{2d}$ for 2D images has been learned, we will use it to learn a point-wise feature extraction function $f^{3d}$ for 3D point clouds. The input to this feature extraction is a point cloud of $L$ points, and the output is a 2D tensor of size $L{\times}C_2$, where $C_2$ is the number of feature dimensions. Each point of the point cloud has a corresponding $C_2$-dimensional feature vector. To learn the feature extraction $f^{3d}$, we develop a novel scheme that uses 2D-to-3D knowledge transfer. We use both global and point-wise knowledge transfer.


\myheading{Global knowledge transfer.} For  global knowledge transfer, we minimize the distance between the aggregated 2D feature vector and the aggregated 3D feature vector by
\begin{align}
    \mL^{3d}_{glb} =\frac{1}{n}\sum_{i=1}^{n}  \left\| g^{2d}(\max_{j} f^{2d}(\y_{ij})) \right.
     \left. - g^{3d}(\max_{}f^{3d}(P_i)) \right\|^2,  \nonumber
\end{align}
where $P_i$ is the point cloud of the $i^{th}$ object and $\y_{ij}$ is the $j^{th}$ view of the $i^{th}$ object. $f^{2d}$ is the feature extractor for 2D images, which was explained in \Sref{sec:step1}. Function $\max_{j} f^{2d}(\y_{ij})$ is the pixel-wise max-pooling across different 2D views. Function $f^{3d}$ is the feature extractor for 3D point cloud, which we seek to learn here. $\max_{} f^{3d}(P_i)$ is element-wise max-pooling among all feature vectors of all points of point cloud $P_i$. Both $g^{2d}$ and $g^{3d}$ are nonlinear projection layers that transform 2D feature and 3D feature vectors to the feature space, respectively.


\myheading{Point-wise knowledge transfer:} 
In addition to global knowledge transfer, 
we use contrastive learning that minimizes the distance between feature representation of a 3D point and its corresponding 2D pixel. 
To determine the point-to-pixel correspondences, we project each point of the point cloud $P_i$ to each image view $\y_{ij}$ using the camera projection matrix $M_{ij}$ to have $\y^{2d}_{ij} = M_{ij} P_i$, 
where $\y^{2d}_{ij}$ is a set of pixels from the rendered image $\y_{ij}$ corresponding to $P_i$. 
For point-wise knowledge transfer, in each optimization iteration, we sample a batch of $k$ corresponding pixel-point pairs, and let $\{(\z_i^{2d}, \z_i^{3d})\}_{i=1}^{k}$ be the corresponding set of feature vector pairs. For the $i^{th}$ pixel-point pair, $\z_i^{2d}$ is obtained by: (1) using the 2D feature extraction function $f^{2d}$ on the image that contains the pixel; (2) passing the output to  the upsampling feature projection module $u^{2d}$; and (3) extracting the feature vector at the location of the pixel in the projected feature map. $\z_i^{3d}$ is obtained by: (1) using the 3D feature extraction function $f^{3d}$ on the point cloud containing that point; (2) passing the output through the projection function $g^{3d}$; and (3) extracting the corresponding feature vector of the point in the point cloud.


For point-wise knowledge transfer, we maximize the similarity between the pixel representation vector and the point representation vector, using a loss function inspired by SimCLR~\cite{chen2020simple}. The loss term for the $i^{th}$ pixel-point pair is:
\begin{align}
\mL^{3d}_{pnt}(i) = &-\log\left(\frac{\fsim(\z^{2d}_i, \z^{3d}_i) }{\fsim(\z^{2d}_i, \z^{3d}_i) + \sum_{j\neq i} (\fsim(\z^{2d}_i, \z^{2d}_j) +  \fsim(\z^{2d}_i, \z^{3d}_j))}\right) \\
     &-\log\left(\frac{\fsim(\z^{3d}_i, \z^{2d}_i) }{\fsim(\z^{3d}_i, \z^{2d}_i) + \sum_{j\neq i} (\fsim(\z^{3d}_i, \z^{2d}_j) + \fsim(\z^{3d}_i, \z^{3d}_j))}\right), \nonumber
\end{align}
where $\fsim(\cdot, \cdot)$ is the exponential cosine function defined in \Eref{eq:sim}.
Intuitively, both 2D and 3D features can be regarded as augmentations of a common latent feature, so they form a positive pair of which similarity can be maximized with the contrastive loss.
The total loss function for a batch of $k$ pixel-point pairs is: 
$    \mL^{3d}_{pnt} = \frac{1}{k}\sum_{i=1}^{k}\mL^{3d}_{pnt}(i)$.

\myheading{Combined loss function.} To pre-train the point cloud network, we minimize a loss function that is the weighted combination of the global knowledge transfer loss and the point-wise knowledge transfer loss: 
\begin{align}
    \mL^{3d} = \lambda_{glb}\mL^{3d}_{glb} + \lambda_{pnt}\mL^{3d}_{pnt}.
\end{align}
In our experiments, we simply use $\lambda_{glb} = \lambda_{pnt}= 1$. 
After pre-training, we can now use the pre-trained weights to initialize the training of downstream tasks. 

\section{Experiments}


\subsection{Implementation details}
\myheading{Dataset for pre-training.} Unless otherwise mentioned, we use ModelNet40 \cite{Wu_2015_CVPR} for pre-training. ModelNet40 is a synthetic dataset that includes 9,480 training samples and 2,468 test samples in 40 categories. ModelNet40 represents each object by a 3D mesh, making it suitable for our multi-view rendering purpose.
For each object in the training set of ModelNet40, we generate its point cloud using farthest-point sampling. 
We render the object into multi-view images by moving a camera around the object. 
Unless otherwise mentioned, each point cloud has 1024 points rendered into 12 views with $32{\times} 32$ resolution.
We use 12 views as they tend to cover all object directions in general. 
We keep the views in low resolution of $32{\times} 32$ to avoid out of memory usage at training.

Our multi-view rendering is implemented as follows. First, each object is normalized into a unit cube. To generate multi-view images from a mesh object, we used Blender~\cite{blender2018} with fixed camera parameters (focal length 35, sensor width 32, and sensor height 32) and a light source. The camera positions are chosen to cover the surrounding views of the object, and the distances from each camera to its neighbor positions are equal.

\myheading{2D feature representation learning.}
We use ResNet50 \cite{he2016deep} as a 2D feature extractor $f^{2d}$ in 2D self-supervised learning process. We use Adam optimizer with the initial learning rate 0.001 without learning decay. We then train the self-supervised model with 1000 epochs and batch size 512.

\myheading{3D feature representation learning.}
We experiment with two common backbones PointNet~\cite{Qi_2017_CVPR} and DGCNN~\cite{wang2019dynamic}, which can be utilized for both classification and segmentation tasks. For PointNet \cite{Qi_2017_CVPR}, we use Adam optimizer with the initial learning rate 0.001, which decays 0.7 every 20 epochs. The momentum of batch normalization starts as 0.5, then divided by 2 every 20 epochs. For DGCNN \cite{wang2019dynamic}, we use an SGD optimizer with the initial learning rate 0.1 and momentum 0.9. We use CosineAnnealingLR scheduler~\cite{loshchilov2016sgdr} for learning rate decay. For both backbones, we train the model with 250 epochs, 200 epochs, and 100 epochs for classification, part segmentation, and semantic segmentation task, respectively, with the same batch size as 32.
After getting the pre-trained models, we test their effectiveness in training with different downstream tasks.

\begin{table}[t!]
\begin{center}
\resizebox{0.8\linewidth}{!}{
\begin{tabular}{l cccccc}
\toprule
    \multirow{2}{*}{} &
      \multicolumn{6}{c}{DGCNN}\\
        \cmidrule{2-7}
    &  Random & Jigsaw & OcCo & CM & STRL & Ours \\
\midrule
MN40 \cite{Wu_2015_CVPR} & 92.7{\scriptsize \textcolor{black}{$\pm$0.1}} & 92.9{\scriptsize \textcolor{black}{$\pm$0.1}} & 92.9{\scriptsize \textcolor{black}{$\pm$0.0}}& 93.0{\scriptsize \textcolor{black}{$\pm$0.1}} & 93.1{\scriptsize \textcolor{black}{$\pm$0.1}} &  \textbf{93.2{\scriptsize \textcolor{black}{$\pm$0.1}}} \\
SO \cite{uy-scanobjectnn-iccv19} & 82.8{\scriptsize \textcolor{black}{$\pm$0.5}}& 82.1{\scriptsize \textcolor{black}{$\pm$0.2}} & 83.2{\scriptsize \textcolor{black}{$\pm$0.2}} & 83.0{\scriptsize \textcolor{black}{$\pm$0.2}} & 83.2{\scriptsize \textcolor{black}{$\pm$0.2}} & \textbf{84.3{\scriptsize \textcolor{black}{$\pm$0.6}}}   \\
SO BG \cite{uy-scanobjectnn-iccv19} & 81.4{\scriptsize \textcolor{black}{$\pm$0.5}} & 82.0{\scriptsize \textcolor{black}{$\pm$0.4}}& 82.9{\scriptsize \textcolor{black}{$\pm$0.4}} & 82.2{\scriptsize \textcolor{black}{$\pm$0.2}} & 83.2{\scriptsize \textcolor{black}{$\pm$0.2}}& \textbf{84.5{\scriptsize \textcolor{black}{$\pm$0.6}}} \\
\bottomrule
\end{tabular}
}
\end{center}
\vskip -0.1in
\caption{Comparison among random, Jigsaw~\cite{sauder2019self}, OcCo~\cite{wang2021unsupervised}, CM~\cite{jing2021self}, STRL~\cite{huang2021spatio}, and our initialization to the object classification downstream task. We reported the mean and standard deviation at the best epoch over three runs.}\label{tab:classification}
\vskip -0.2in
\end{table}

\subsection{Object classification}

We first experiment with object classification for 3D point cloud analysis. Two standard benchmarks are used, namely ModelNet40 \cite{Wu_2015_CVPR} and ScanObjectNN \cite{uy-scanobjectnn-iccv19} dataset. 
We follow the previous paper~\cite{wang2021unsupervised} to use ModelNet40 in both pre-training and downstream tasks. 
ScanObjectNN is a real-world dataset that has 15 categories with 2,321 and 581 samples for training and testing, respectively. We use the default variant (OBJ\_ONLY, denoted by ScanObjectNN) and the variant with background (OBJ\_BG, denoted by ScanObjectNN BG).
We follow the experimental setting in the original PointNet~\cite{Qi_2017_CVPR}.

We compare the performance of DGCNN~\cite{wang2019dynamic} with and without  pre-training. The results are shown in \Tref{tab:classification}. We also provide comparisons with the PointNet backbone~\cite{Qi_2017_CVPR} in the supplementary material.
As can be seen, models with pre-training outperform their randomly initialized counterparts. 
We further compare our method to previous point cloud pre-training methods, including Jigsaw~\cite{sauder2019self}, OcCo~\cite{wang2021unsupervised}, CM~\cite{jing2021self}, and STRL~\cite{huang2021spatio}.
Particularly, Jigsaw~\cite{sauder2019self} learns to solve jigsaw puzzles as a pretext task for pre-training. OcCo~\cite{wang2021unsupervised} is based on mask-based pre-training from natural language processing to propose a point cloud completion task for pre-training.
CM~\cite{jing2021self} considered self-supervision from cross-modality and cross-view feature learning, which shares some similarity to ours. 
Our method differs in that we use a contrastive loss to learn 2D features and an L2 loss to match 2D-3D features while CM~\cite{jing2021self} uses a triplet loss for 2D features and a cross-entropy loss for matching 2D-3D features.
STRL~\cite{huang2021spatio} explored self-supervision with spatial-temporal representation learning.
In \Tref{tab:classification}, it can be seen that our proposed self-supervision with contrastive loss and multi-view rendering outperforms other initialization methods on both ModelNet40 and ScanObjectNN dataset. 


\subsection{Network Analysis}
\begin{figure}[t!]
\begin{subfigure}{.23\textwidth}
  \centering
  \includegraphics[width=1\linewidth]{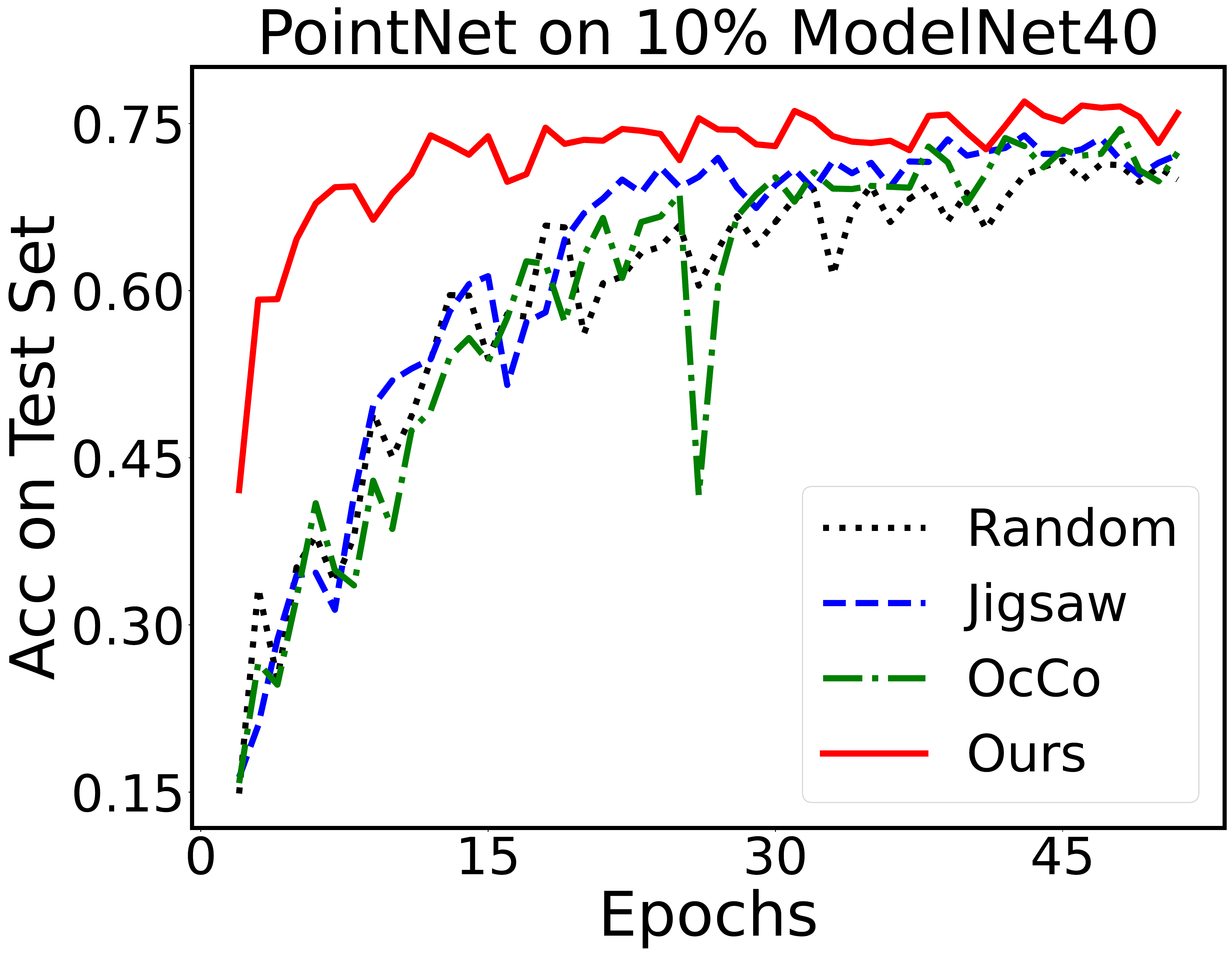}  
\end{subfigure}
~~
\begin{subfigure}{.23\textwidth}
  \centering
  \includegraphics[width=1\linewidth]{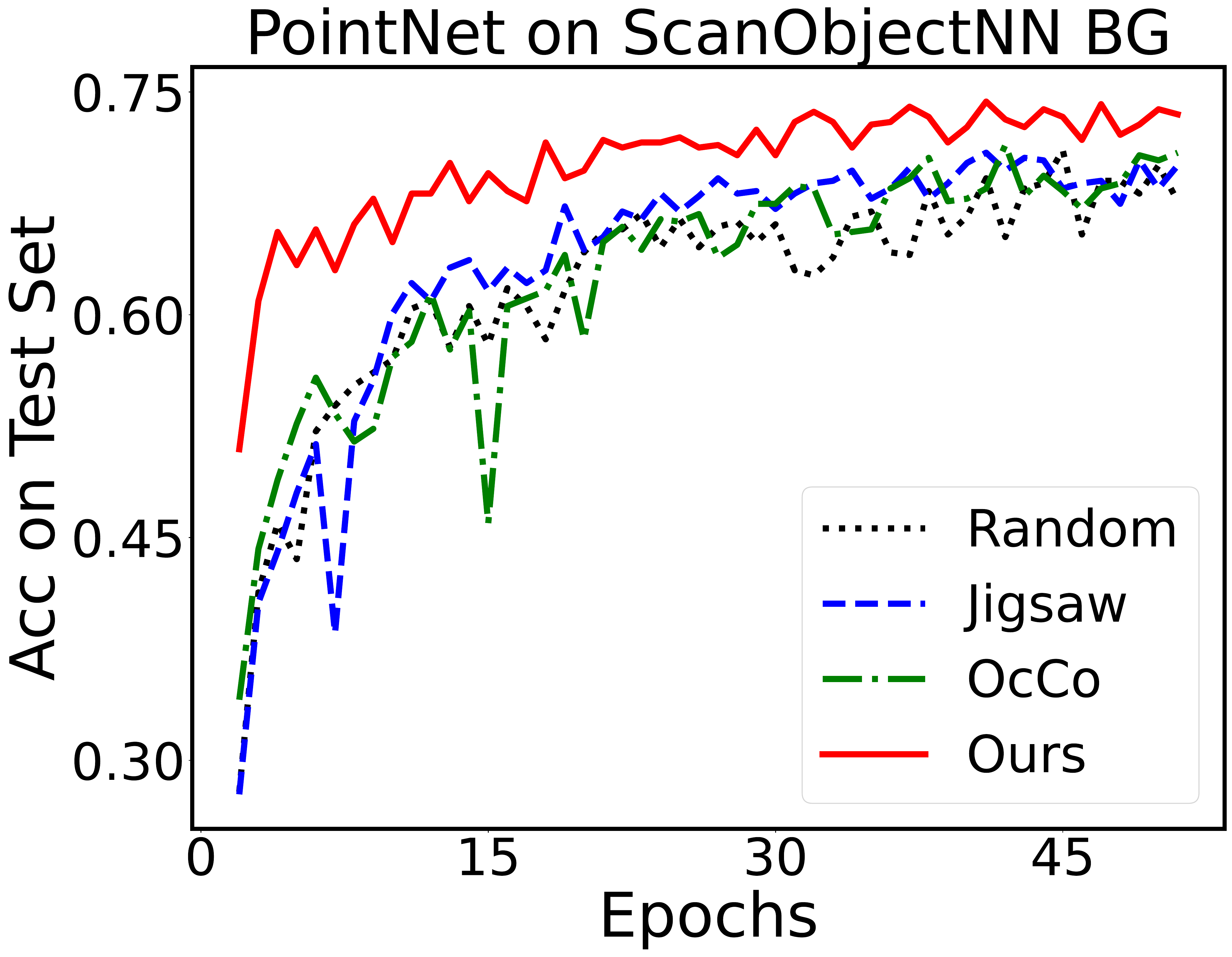}  
\end{subfigure}
~~
\begin{subfigure}{.23\textwidth}
  \centering
  \includegraphics[width=1\linewidth]{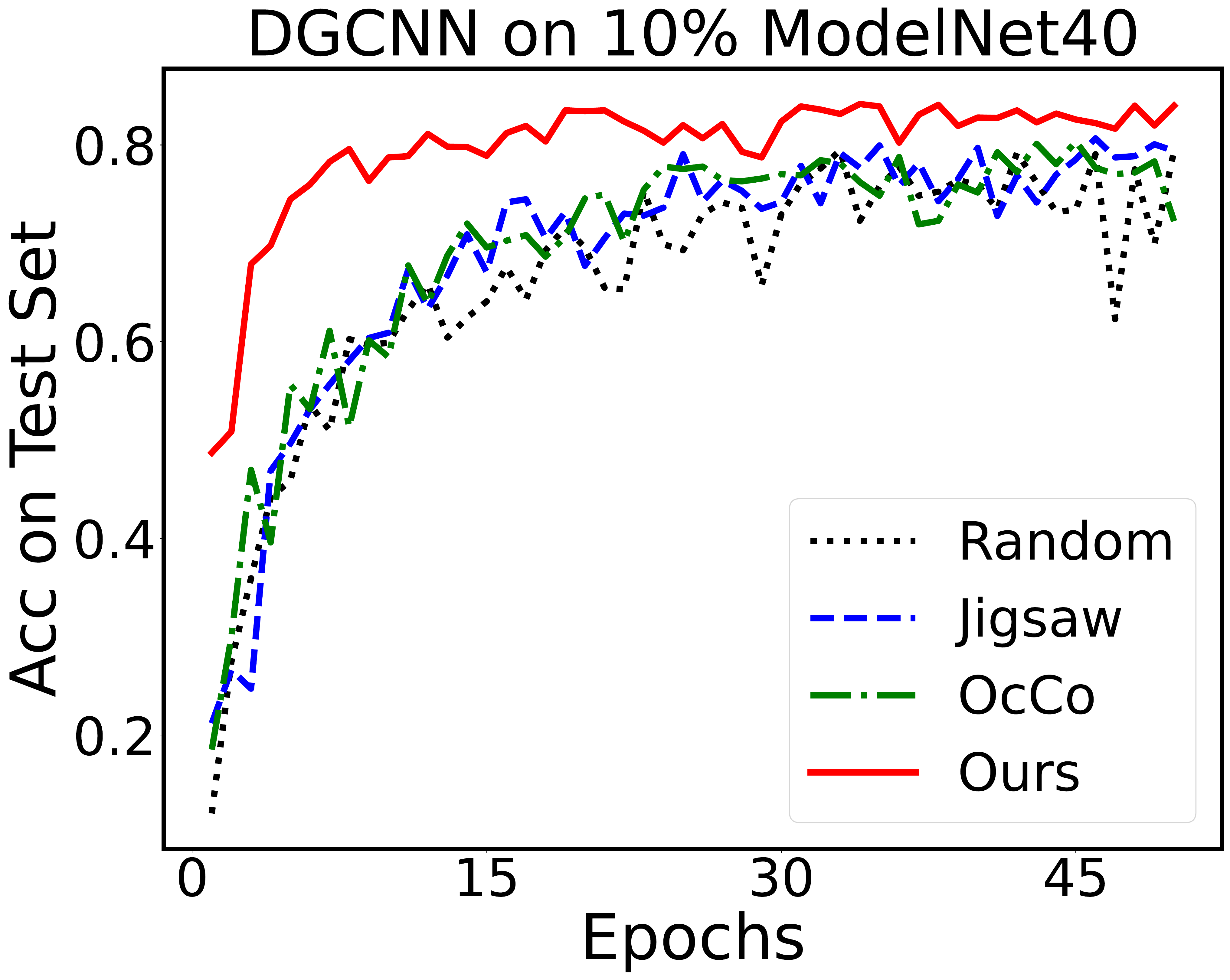}  
\end{subfigure}
~~
\begin{subfigure}{.23\textwidth}
  \centering
  \includegraphics[width=1\linewidth]{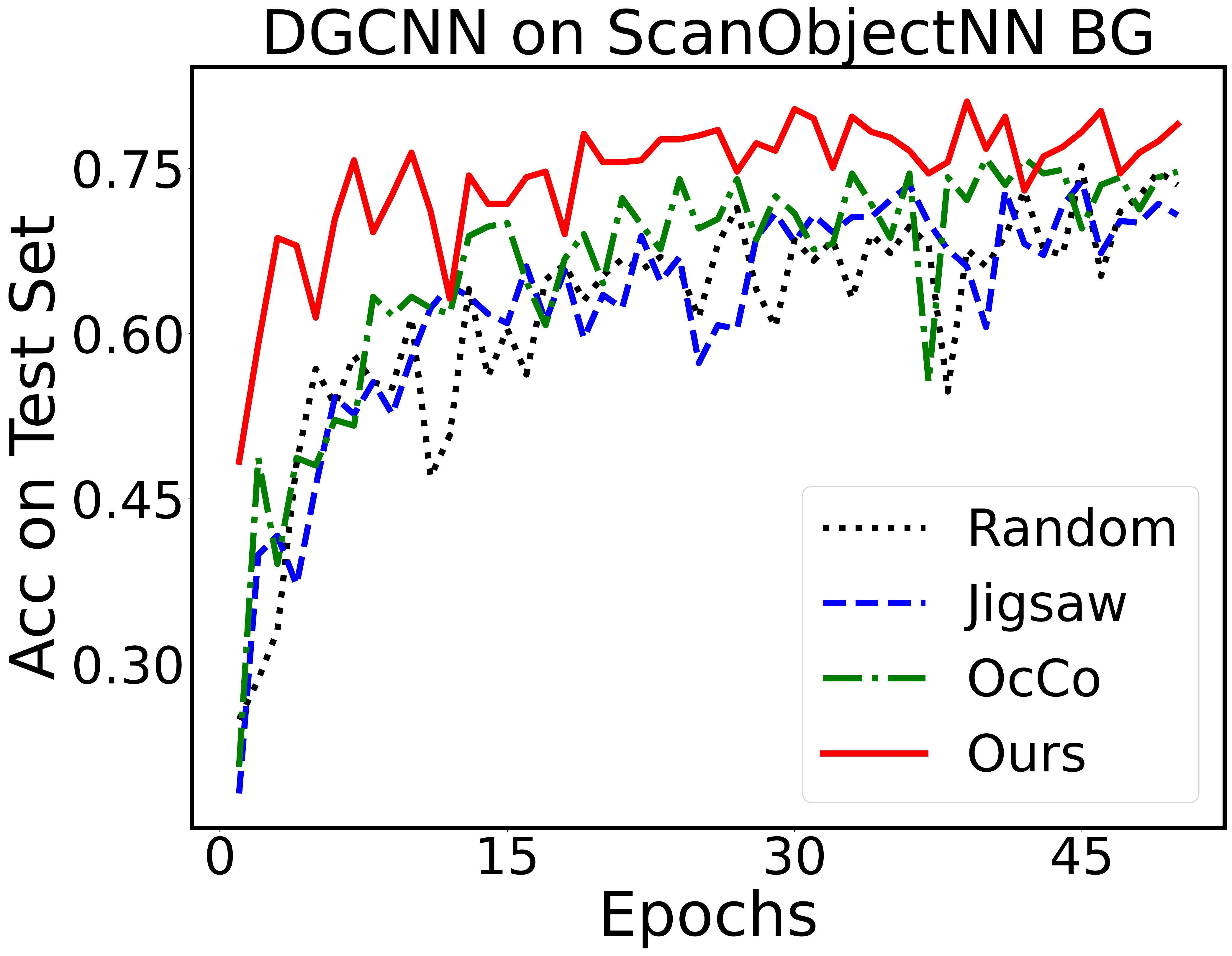}  
\end{subfigure}
\vskip -0.1in
\caption{Test-set accuracy over different training epochs in the object classification task. The plots show that previous pre-training methods are only marginally more effective than random initialization while our method shows a clear improvement. }
\label{fig:learning_curve}
\vskip -0.1in
\end{figure}

\setlength{\tabcolsep}{2pt}
\begin{table}[t]
\small
\begin{center}
\begin{tabular}{l cccc|cccc}
\toprule
    \multirow{2}{*}{} &
      \multicolumn{4}{c|}{PointNet} &
      \multicolumn{4}{c}{DGCNN}\\
        \cmidrule{2-9}
    & Random & Jigsaw & OcCo & Ours & Random & Jigsaw & OcCo & Ours \\
\midrule
5\% & 73.2 & 73.8 & 73.9 & \textbf{77.9} & 82.0 & 82.1 & 82.3 & \textbf{84.9} \\
10\% & 75.2 & 77.3 & 75.6 & \textbf{79.0} & 84.7 & 84.1 & 84.9 & \textbf{86.6} \\
20\% & 81.3 & 82.9 & 81.6 & \textbf{84.6} & 89.4 & 89.2 & 89.1 & \textbf{90.2} \\
50\% & 86.6 & 86.5 & 87.1 & \textbf{87.6} & 91.6 & 91.8 & 91.7 & \textbf{92.4} \\
70 \% & 88.3 & 88.4 & 88.4 & \textbf{88.7} & 92.3 & 92.4 & 92.5 & \textbf{92.8} \\
90 \% & 88.5 & 88.8 & 88.8 & \textbf{89.4} & 92.6 & 92.9 & 92.9 & \textbf{93.1} \\
\bottomrule
\end{tabular}
\end{center}
\vskip -0.1in
\caption{Performance of the object classification task trained with fewer data. Our method has significant gains compared to other initialization methods. We reported the mean at the best epoch over three runs.}\label{tab:ratio}
\vskip -0.3in
\end{table}

\myheading{Accuracy over epochs.} \Fref{fig:learning_curve} plots the accuracy on the test set over different training epochs. The proposed initialization helps both PointNet and DGCNN converge faster and obtain better accuracy than other initialization methods. 
For example, when we use ModelNet40 with 10\% training dataset, the model with our initialization converges after around 15 epochs, while with other initialization methods, it takes around 40 epochs. For ScanObjectNN (OBJ\_BG variant), the models converge after about 20 epochs with our initialization and about 45 epochs with other methods.

\myheading{Training size.} Our pre-training allows the point cloud network to be trained with less data compared to initialization with random weights. To demonstrate this, we reduce the number of samples in the training set of ModelNet40 to 5\%, 10\%, 20\%, 50\%, and 70\%. We then use these datasets to train the model for the object classification task. Finally, we evaluate these learned models on the test set of ModelNet40. 
\Tref{tab:ratio} shows the results with random initialization, Jigsaw~\cite{sauder2019self}, OcCo~\cite{wang2021unsupervised}, and our initialization, respectively. 
As can be seen, models using our proposed initialization outperform other models.

\myheading{Number of views.} We analyze the influence of multi-view rendering in our pre-training performance. 
We render the shapes with 4, 8, 12, and 24 views in object classification task. 
The results are shown in \Tref{tab:num_view}. 
For PointNet, the performance is best with 8 views, while for DGCNN it is generally enough to use 4 views, but DGCNN for ScanObjectNN performs best with 24 views.

\begin{table*}
\setlength\tabcolsep{6pt}
\begin{center}
\resizebox{\linewidth}{!}{
\begin{tabular}{l cccc|cccc}
\toprule
    \multirow{2}{*}{} &
      \multicolumn{4}{c|}{PointNet} &
      \multicolumn{4}{c}{DGCNN}\\
        \cmidrule{2-9}
    & 4-views & 8-views & 12-views & 24-views & 4-views & 8-views & 12-views & 24-views \\
\midrule
MN40 \cite{Wu_2015_CVPR} & 88.9 & \textbf{89.2} & 88.9 & 88.9 & \textbf{92.8} &  92.3 & 92.5 & 92.3\\
SO \cite{uy-scanobjectnn-iccv19} & 79.0 & \textbf{80.4} & 79.3 & 79.1 & 82.7 &  82.6 & 82.8 & \textbf{84.9}\\
SO BG \cite{uy-scanobjectnn-iccv19} & 74.2 & \textbf{77.1} & 75.7 & 76.6 & \textbf{82.8} &  81.9 & 82.6 & 81.4\\
\bottomrule
\end{tabular}
}
\end{center}
\vskip -0.1in
\caption{Performance of object classification tested with different number of views in multi-view rendering.}\label{tab:num_view}
\vskip -0.3in
\end{table*}

\myheading{Classification with SVM.} 
To evaluate the generalization ability of our pre-trained model, we follow a similar test scenario in \cite{sauder2019self}. 
First, we freeze the weights of the pre-trained model and pass the 3D object through this model to obtain their embeddings. Then, we train a Support Vector Machine (SVM) on the embeddings of the train set and evaluate it on the test set on three datasets ModelNet40, ScanObjectNN, and ScanObjectNN (OBJ\_BG variant). For SVM, we used grid search to find the best parameter of SVM with a Radial Basis Function kernel. The results are shown in \Tref{tab:svm}. The proposed initialization outperforms the other initialization methods, Jigsaw and OcCo, sometimes by a wide margin as in the ScanObjectNN (OBJ\_BG variant).
We also provide additional comparisons to previous self-supervised methods on ModelNet40 in \Tref{tab:svm_selfsup}. As can be seen, our proposed method outperforms almost other methods in both PointNet and DGCNN, except in PointNet, our method is ranked second while ParAE~\cite{eckart2021self} performs best.

\begin{table}
\small
\setlength\tabcolsep{6pt}
\begin{center}
\begin{tabular}{l ccc|ccc}
\toprule
    \multirow{2}{*}{} &
      \multicolumn{3}{c|}{PointNet} &
      \multicolumn{3}{c}{DGCNN}\\
        \cmidrule{2-7}
    & Jigsaw & OcCo & Ours & Jigsaw & OcCo & Ours \\
\midrule
ModelNet40 \cite{Wu_2015_CVPR} & 82.5 & 87.2 & \textbf{89.7} & 83.1 &  89.5 & \textbf{91.7} \\
ScanObjectNN \cite{uy-scanobjectnn-iccv19}& 49.7 & 62.1 & \textbf{70.2} & 57.8 & 69.0 & \textbf{76.3}  \\
ScanObjectNN BG \cite{uy-scanobjectnn-iccv19}& 48.9 & 61.7 & \textbf{69.5} & 51.1 & 67.5 & \textbf{74.2}  \\
\bottomrule
\end{tabular}
\end{center}
\vskip -0.1in
\caption{The result of SVM applied on the object embedding learned from different initializations. It shows that features learned by our method are more discriminative than other methods, resulting in more accurate classifications.}\label{tab:svm}
\vskip -0.3in
\end{table}

\begin{table}[h!]
\small
\setlength\tabcolsep{6pt}
\begin{center}
\begin{tabular}{l cc}
\toprule
    & PointNet & DGCNN   \\
\midrule
Rotation \cite{poursaeed2020self} & 88.6 & 99.8 \\
STRL \cite{huang2021spatio} & 88.3 & 90.9 \\
ParAE \cite{eckart2021self} & \textbf{90.3} & 91.6 \\
CrossPoint \cite{afham2022crosspoint} & 89.1 & 91.2 \\
Ours & 89.7 & \textbf{91.7} \\
\bottomrule
\end{tabular}
\end{center}
\vskip -0.1in
\caption{More comparisons of SVM classification on ModelNet40.}\label{tab:svm_selfsup}
\vskip -0.2in
\end{table}

\begin{table}[h!]
\small
\setlength\tabcolsep{6pt}
\begin{center}
\begin{tabular}{l ccc|ccc}
\toprule
    \multirow{2}{*}{} &
      \multicolumn{3}{c|}{PointNet} &
      \multicolumn{3}{c}{DGCNN}\\
        \cmidrule{2-7}
    & $\mL^{3d}_{glb}$ & $\mL^{3d}_{pnt}$ & $\mL^{3d}$ & $\mL^{3d}_{glb}$ & $\mL^{3d}_{pnt}$ & $\mL^{3d}$ \\
\midrule
ModelNet40 \cite{Wu_2015_CVPR} & 88.5 & 88.5 & \textbf{88.9} &  92.4 & 92.1 & \textbf{92.5} \\
ScanObjectNN \cite{uy-scanobjectnn-iccv19} & 77.6 & 78.8 & \textbf{79.3}  &  81.8 & 81.1 & \textbf{82.8} \\
ScanObjectNN BG \cite{uy-scanobjectnn-iccv19} & 74.5 & 74.2 & \textbf{75.7} & 81.6 & 81.6 & \textbf{82.6} \\
\bottomrule
\end{tabular}
\end{center}
\vskip -0.1in
\caption{Effect of loss function choice to our pre-training.}\label{tab:loss}
\vskip -0.2in
\end{table}

\begin{table}[h!]
\setlength\tabcolsep{6pt}
\begin{center}
\begin{tabular}{l ccc}
\toprule
    & RGB & Silhouette & Shading  \\
\midrule
ModelNet40 \cite{Wu_2015_CVPR} & 88.3 & \textbf{88.9} & \textbf{88.9}  \\
ScanObjectNN \cite{uy-scanobjectnn-iccv19} & \textbf{79.7} & 78.8 & 79.3  \\
ScanObjectNN BG \cite{uy-scanobjectnn-iccv19} & 75.1 & 75.6 & \textbf{75.7}  \\
\bottomrule
\end{tabular}
\end{center}
\vskip -0.1in
\caption{Effect of rendering techniques to the pre-training on PointNet~\cite{Qi_2017_CVPR}.}\label{tab:image}
\vskip -0.4in
\end{table}

\myheading{Ablation study of loss functions.} \Tref{tab:loss} reports the performance of our method for the classification task when both global loss and pixel-point loss are used together or individually. 
The network achieves the best performance when trained with both losses. 
Using either global loss or pixel-point loss results in accuracy drop especially for the ScanObjectNN dataset~\cite{uy-scanobjectnn-iccv19}.
This is because the global loss is only useful in distilling knowledge from an image view to a point cloud while the pixel-point loss encourages the model learn consistent features locally. Using both losses lets the model have the best of both worlds.

\begin{figure}[h!]
\centering
\vskip -0.2in
\includegraphics[width=0.15\linewidth]{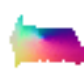} 
\includegraphics[width=0.15\linewidth]{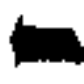}
\quad
\includegraphics[width=0.15\linewidth]{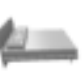}
\vskip -0.2in
\caption{Object rendering with position encoding (RGB), silhouette, and shading.}
\label{fig:type_image}
\end{figure}

\myheading{Multi-view rendering.} Our pre-training requires multi-view image rendering, which can be implemented by a wide range of rendering techniques. 
We study the effect of images rendered from the object mesh, 3D position encoding, and object silhouette on the classification task (please refer to example rendering in \Fref{fig:type_image}).
For the original object mesh, we use Blender~\cite{blender2018} to render the object geometry with Phong shading, resulting in grayscale \emph{shaded} images.
For 3D position encoding, the images are rendered directly from point clouds. Specifically, we first assign a pseudo-color (RGB) to each point of a point cloud based on their 3D coordinates, then project the points to the image plane with preset camera projection matrices.
For object silhouette, the process is generally similar except that we use black instead of the pseudo-color for each point in the point clouds. 
For pixel that has more than one corresponding point, we choose the point with minimum distance to the camera.  

The performance of object classification with different rendering techniques is reported in \Tref{tab:image}, where there is no best technique overall. 
Using shaded images results in slightly higher accuracy than using position encoding and silhouette rendering in ModelNet40 and ScanObjectNN BG. This is because shaded images often have more details than position encoding and silhouette images since shaded images are rendered from meshes. 
Exploring more robust rendering techniques for self-supervised learning is left for future work.


\subsection{Part segmentation and scene segmentation}

\begin{table*}[t!]
\setlength\tabcolsep{5pt}
\begin{center}
\resizebox{\linewidth}{!}{
\begin{tabular}{l cccc|cccc}
\toprule
    \multirow{2}{*}{} &
      \multicolumn{4}{c|}{PointNet} &
      \multicolumn{4}{c}{DGCNN}\\
        \cmidrule{2-9}
    & Random & Jigsaw & OcCo & Ours & Random & Jigsaw & OcCo & Ours \\
\midrule
mAcc & 93.3 {\scriptsize \textcolor{black}{$\pm$0.2}} & 93.0 {\scriptsize \textcolor{black}{$\pm$0.0}} & 93.3 {\scriptsize \textcolor{black}{$\pm$0.1}} & \textbf{93.4 {\scriptsize \textcolor{black}{$\pm$0.0}}} & 94.2 {\scriptsize \textcolor{black}{$\pm$0.0}} &  94.1 {\scriptsize \textcolor{black}{$\pm$0.0}} & \textbf{94.3 {\scriptsize \textcolor{black}{$\pm$0.0}}} & 94.2 {\scriptsize \textcolor{black}{$\pm$0.1}} \\
mIoU & 83.1 {\scriptsize \textcolor{black}{$\pm$0.3}} & 83.2 {\scriptsize \textcolor{black}{$\pm$0.1}} & 83.0 {\scriptsize \textcolor{black}{$\pm$0.2}} & \textbf{83.3 {\scriptsize \textcolor{black}{$\pm$0.1}}} & \textbf{84.7 {\scriptsize \textcolor{black}{$\pm$0.0}}} & 84.5 {\scriptsize \textcolor{black}{$\pm$0.1}} & \textbf{84.7 {\scriptsize \textcolor{black}{$\pm$0.1}}} & \textbf{84.7 {\scriptsize \textcolor{black}{$\pm$0.1}}} \\
\bottomrule
\end{tabular}
}
\end{center}
\vskip -0.1in
\caption{The result of four initialization in the part segmentation task on the ShapeNetPart dataset \cite{wang2019dynamic}. We reported the mean and standard error of mAcc and mIoU over three runs.}\label{tab:parseg}
\vskip -0.3in
\end{table*}

\begin{table*}[t!]
\small
\setlength\tabcolsep{6pt}
\begin{center}
\begin{tabular}{c cccc|cccc}
\toprule
    \multirow{2}{*}{} &
      \multicolumn{4}{c|}{PointNet} &
      \multicolumn{4}{c}{DGCNN}\\
        \cmidrule{2-9}
    & Random & Jigsaw & OcCo & Ours & Random & Jigsaw & OcCo & Ours \\
\midrule
mAcc & 83.9 & 82.5 & 83.6 & \textbf{85.0} & 86.8 &  86.8 & \textbf{87.0} & \textbf{87.0} \\
mIoU & 43.6 & 43.6 & 44.5 & \textbf{46.7} & 49.3 & 48.2 & 49.5 & \textbf{49.9} \\
\bottomrule
\end{tabular}
\end{center}
\vskip -0.1in
\caption{Fold 1 of overall point accuracy (mAcc) and mean Intersection-over-Union (mIoU) on the S3DIS (Stanford Area 5 Test) \cite{armeni20163d}.}
\label{tab:seemseg}
\vskip -0.3in
\end{table*}

Beyond classification, we conduct experiments to validate our pre-training for semantic part segmentation and scene segmentation tasks. 
We first experiment with object part segmentation on the ShapeNetPart dataset~\cite{wang2019dynamic} that includes 16,881 objects from 16 categories. Each object is represented by 2,048 points, and each point belongs to one of 50 part types. Most objects in the dataset are divided into two to five parts. 
As shown in \Tref{tab:parseg}, our initialization is slightly better than random initialization, Jigsaw, and OcCo in both overall point accuracy (mAcc) and mean Intersection-over-Union (mIoU) metric. We observed that the improvement is minor in the part segmentation task because the network architecture used for part segmentation is largely different from the pre-trained networks, and therefore only a few layers of the part segmentation networks can be initialized by the pre-trained model.


We also experiment with semantic scene segmentation on the Stanford Large-Scale 3D Indoor Spaces dataset (S3DIS)~\cite{armeni20163d}. 
This dataset contains point clouds of 272 rooms from 6 areas and 13 categories. Each room is split into $1m {\times} 1m$ blocks. Each point is represented by a 9D vector including XYZ, RGB, and normalized location in the room. During training, each block is sampled with 4096 points, but during testing, all points are used. 
The results are shown in \Tref{tab:seemseg}. 
We see that models initialized by our method outperform others in both PointNet~\cite{Qi_2017_CVPR} and DGCNN~\cite{wang2019dynamic}.

\subsection{Pre-training with synthetic vs. real-world data}
Multi-view rendering can be easily used for self-supervised learning when working with synthetic data as we have shown with ModelNet40~\cite{Wu_2015_CVPR}. Real-world 3D datasets, however, often do not provide such multi-view images, limiting our choices for pre-training. In this section, we investigate the role of synthetic and real-world data in pre-training by comparing to PointContrast~\cite{xie2020pointcontrast} and DepthContrast~\cite{zhang2021self} on the segmentation and detection task.
We run different experiments using Sparse Residual U-Net (SR-UNet)~\cite{choy20194d} as the network backbone. 
Compared to PointNet and DGCNN backbone used in the previous sections, SR-UNet uses sparse convolutions to learn features on point clouds, discarding the need of a sliding window for processing large-scale point clouds.
For pre-training data, we use ModelNet40~\cite{Wu_2015_CVPR} as synthetic data and ScanNet~\cite{dai2017scannet} as real data. 
 
 \myheading{Pre-training.} As the original PointContrast~\cite{xie2020pointcontrast} only supports ScanNet for pre-training, here we adapt ModelNet40 to PointContrast by using surface point cloud pairs, formed for every two continuous views, instead of the provided point cloud pairs from ScanNet. As for our model, we use two view images when their corresponding point cloud pairs have at least 30\% overlapping. To define pixel-point pairs, we reconstruct a point cloud from the first depth image in an image pair, then project it to two color images to get pixel-point correspondences. 
 During training, we follow original setting of PointContrast~\cite{xie2020pointcontrast}.
 For our pre-trained model on ScanNet~\cite{dai2017scannet}, we used the pre-trained ResNet50~\cite{he2016deep} on ImageNet provided by Pytorch\cite{NEURIPS2019_9015} as the 2D feature extractor. Besides, all images are resized to $240 \times 320$ and we only use the point-wise knowledge transfer loss for pre-training.
 We train the model with one GPU and four GPUs for ModelNet40 and ScanNet datasets, respectively.
 
 \setlength\tabcolsep{5pt}
\begin{table*}[t!]
\begin{center}
\resizebox{\linewidth}{!}{
\begin{tabular}{ccccccc}
\toprule
 Dataset & \stackbox{Task\\(Metric)} &  \stackbox{Random\\~} & \stackbox{PC~\cite{xie2020pointcontrast}\\ModelNet} & \stackbox{Ours\\ModelNet} & \stackbox{PC~\cite{xie2020pointcontrast}\\ScanNet} & \stackbox{Ours\\ScanNet}  \\
\midrule
S3DIS (Area 5) & sem. seg. (Acc) & 72.5 &  71.2 {\scriptsize \textcolor{gray}{-1.3}} & \textbf{73.2} {\scriptsize \bf\textcolor{teal}{+0.7}} & 73.0 {\scriptsize \textcolor{gray}{+0.5}}& 73.0 {\scriptsize \textcolor{gray}{+0.5}}\\
S3DIS (Area 5) & sem. seg. (IoU) & 64.5  & 64.1 {\scriptsize \textcolor{gray}{-0.4}} & 66.0 {\scriptsize \textcolor{gray}{+1.5}} & 66.1 {\scriptsize \textcolor{gray}{+1.6}} & \textbf{66.5} {\scriptsize \bf\textcolor{teal}{+2.0}}\\
\midrule
ScanNet & sem. seg. (Acc) & 80.2 & 80.3 {\scriptsize \textcolor{gray}{+0.1}}& \textbf{81.1} {\scriptsize \bf\textcolor{teal}{+0.9}} & 80.8 {\scriptsize \textcolor{gray}{+0.6}} & 81.0 {\scriptsize \textcolor{gray}{+0.8}}\\
ScanNet & sem. seg. (IoU) & 72.4 & 72.5 {\scriptsize \textcolor{gray}{+0.1}} & 73.3 {\scriptsize \textcolor{gray}{+0.9}} & 73.1 {\scriptsize \textcolor{gray}{+0.7}} & \textbf{73.6} {\scriptsize \bf\textcolor{teal}{+1.2}}\\
\midrule
ScanNet & 3D det. ($\mbox{AP}_{50}$) & 35.2 & 36.6 {\scriptsize \textcolor{gray}{+1.4}} & 38.2 {\scriptsize \textcolor{gray}{+3.0}} & 36.1 {\scriptsize \textcolor{gray}{+0.9}} & \textbf{39.2} {\scriptsize \bf\textcolor{teal}{+4.0}} \\
ScanNet & 3D det. ($\mbox{AP}_{25}$) & 56.5 & 58.2 {\scriptsize \textcolor{gray}{+1.7}} & 58.4 {\scriptsize \textcolor{gray}{+1.9}} & 59.5 {\scriptsize \textcolor{gray}{+3.0}} & \textbf{60.3} {\scriptsize \bf\textcolor{teal}{+3.8}} \\
\midrule
SUN RGB-D & 3D det. ($\mbox{AP}_{50}$) & 32.3 & 34.8 {\scriptsize \textcolor{gray}{+2.5}} & 34.9 {\scriptsize \textcolor{gray}{+2.6}} & 34.8 {\scriptsize \textcolor{gray}{+2.5}} & \textbf{35.1} {\scriptsize \bf\textcolor{teal}{+2.8}} \\
SUN RGB-D & 3D det. ($\mbox{AP}_{25}$) & 55.5 & 57.8 {\scriptsize \textcolor{gray}{+2.3}} & \textbf{58.1} {\scriptsize \bf\textcolor{teal}{+2.6}} & 57.4 {\scriptsize \textcolor{gray}{+1.9}} & 57.8 {\scriptsize \textcolor{gray}{+2.3}} \\
\bottomrule
\end{tabular}}
\end{center}
\vskip -0.1in
\caption{Comparison to PointContrast (PC) ~\cite{xie2020pointcontrast} on the semantic segmentation and 3D object detection task on S3DIS dataset~\cite{armeni20163d}, ScanNet dataset~\cite{dai2017scannet}, and SUN RGB-D dataset~\cite{song2015sun}. Our method outperforms PointContrast when pre-trained on both datasets. The subscript indicates the performance difference compared to the Random case.}\label{tab:downstream}
\vskip -0.2in
\end{table*}

 \myheading{Segmentation and detection results.}
 We evaluate four pre-training configurations with the semantic segmentation task on two datasets S3DIS~\cite{armeni20163d} and ScanNet~\cite{dai2017scannet}. We show the results in \Tref{tab:downstream} (comparisons to PointContrast~\cite{xie2020pointcontrast}) and \Tref{tab:downstream_depthcontrast} (comparisons to DepthContrast~\cite{zhang2021self}). In \Tref{tab:downstream}, on both datasets, the performance gap between our models pre-trained on synthetic and real data is small. When testing on S3DIS, our pre-trained network on ModelNet even provides a slightly better performance compared to the pre-trained model on ScanNet on Acc. metric, and it offers $2\%$ increase when compared with the PointContrast counterpart on both Acc. and IoU metric.
 \setlength\tabcolsep{5pt}
\begin{table*}[h!]
\small
\begin{center}
\resizebox{\linewidth}{!}{
\begin{tabular}{cccccc}
\toprule
 Dataset & \stackbox{Task\\(Metric)} &  \stackbox{Random\\~} & \stackbox{DepthContrast~\cite{zhang2021self}\\~} & \stackbox{Ours\\ModelNet} &\stackbox{Ours\\ScanNet}  \\
\midrule
S3DIS (Area 5) & sem. seg. (Acc) & 70.9 &  72.1 {\scriptsize \textcolor{gray}{+1.2}} & \textbf{75.1} {\scriptsize \bf\textcolor{teal}{+4.2}} & 74.5 {\scriptsize \textcolor{gray}{+3.6}}\\
S3DIS (Area 5) & sem. seg. (IoU) & 64.0  & 64.8 {\scriptsize \textcolor{gray}{+0.8}} &  \textbf{66.8} {\scriptsize \bf\textcolor{teal}{+2.8}} & 66.5 {\scriptsize \textcolor{gray}{+2.5}}\\
\midrule
ScanNet & sem. seg. (Acc) & 77.2 & 77.6 {\scriptsize \bf\textcolor{gray}{+0.4}}& 77.4 {\scriptsize \textcolor{gray}{+0.2}} & \textbf{78.3} {\scriptsize \bf\textcolor{teal}{+1.1}}\\
ScanNet & sem. seg. (IoU) & 69.1 & 69.9 {\scriptsize \bf\textcolor{gray}{+0.8}} & 69.2 {\scriptsize \textcolor{gray}{+0.1}} & \textbf{70.7} {\scriptsize \bf\textcolor{teal}{+1.6}}\\
\bottomrule
\end{tabular}
}
\end{center}
\vskip -0.1in
\caption{Comparison to DepthContrast~\cite{zhang2021self} on the semantic segmentation task on S3DIS dataset~\cite{armeni20163d} and ScanNet dataset~\cite{dai2017scannet}. The subscript indicates the performance difference compared to the Random case.}\label{tab:downstream_depthcontrast}
\vskip -0.3in
\end{table*}
 In \Tref{tab:downstream_depthcontrast}, we also compare with DepthContrast on semantic segmentation task. For S3DIS, our pre-trained models on both synthetic and real data achieve better performance approximately $2\%$ on IoU and $4\%$ on Acc. For ScanNet, our pre-trained model on synthetic data outperforms the random setting but is slightly less effective than DepthContrast. However, our pre-trained model on real data outperforms both random and DepthContrast initialization.

 
 We also perform comparison on the 3D object detection task on the ScanNet dataset~\cite{dai2017scannet} and SUN RGB-D dataset~\cite{song2015sun}. 
 Following \cite{xie2020pointcontrast}, we replace original PointNet++~\cite{qi2017pointnet++} backbone of VoteNet~\cite{qi2019deep} by SR-UNet~\cite{choy20194d}. The results are also shown in \Tref{tab:downstream}. As can be seen, our method outperforms PointContrast when pre-training on the same dataset. When using synthetic data, our model can obtain two points higher in $mAP_{50}$ compared with the PointContrast counterpart. When using real data, the $mAP$ scores increase slightly and achieve state-of-the-art performance.
 


\section{Conclusion}
We propose a self-supervised learning method based on multi-view rendering to pre-train 3D point cloud neural networks. Our self-supervision with multi-view rendering on global and local loss functions yield state-of-the-art performance on several downstream tasks including object classification, semantic segmentation and object detection. Our pre-training method works well on both synthetic and real-world data; it also proves the effectiveness of pre-training on synthetic data like ModelNet40 for downstream tasks with real data like semantic segmentation and 3D object detection.  

%
%

%
%
%
\bibliographystyle{splncs04}
\bibliography{egbib}

\newpage
\appendix
\begin{center}
\textbf{\Large{Supplementary Material}}
\end{center}

In this document, we provide more details for our proposed method. Firstly, we present the complete experiment of using different rendering styles, including position encoding, silhouette, and shading for both PointNet and DGCNN (Section~\ref{sec:render}).
We also include additional results for rest of object classification task with PointNet backbone(Section~\ref{sec:cls}), 6-fold cross validation results for the semantic segmentation task on S3DIS dataset (Section~\ref{sec:semseg}).
Secondly, we also report the performance of training with limited data on both ModelNet40 and ScanObjectNN (Section~\ref{sec:limited}).
Finally, we report detail settings, runtime statistics and more insights into the proposed method by analyzing the t-SNE embedding and the critical 

\section{Evaluation of multi-view rendering}
\label{sec:render}
We report the performance of PointNet~\cite{Qi_2017_CVPR} and DGCNN~\cite{wang2019dynamic} on different rendering styles in Table \ref{tab:image}. It can be seen that shaded images yield slightly higher performance than other renderings on both datasets. 
However, other rendering styles such as position encoding (RGB) and silhouette still produce competitive results. 
It implies that in cases where only point clouds are available for pre-training, RGB and silhouette rendering can be used while not causing a significant performance difference compared to mesh-based rendering.  
\begin{table}[h!]
\begin{center}
\resizebox{\linewidth}{!}{
\begin{tabular}{l ccc|ccc}
\toprule
    \multirow{2}{*}{} &
      \multicolumn{3}{c|}{PointNet} &
      \multicolumn{3}{c}{DGCNN}\\
        \cmidrule{2-7}
    & RGB & Silhouette & Shading & RGB & Silhouette & Shading  \\
\midrule
ModelNet40 \cite{Wu_2015_CVPR} & 88.3 $\pm$ 0.2 & \textbf{88.9 $\pm$ 0.2} & \textbf{88.9 $\pm$ 0.1} &  \textbf{92.5 $\pm$ 0.2} & \textbf{92.5 $\pm$ 0.2} & \textbf{92.5 $\pm$ 0.1} \\
ScanObjectNN \cite{uy-scanobjectnn-iccv19} & \textbf{79.7 $\pm$ 0.5} & 78.8 $\pm$ 0.6 & 79.3 $\pm$ 0.3& \textbf{82.8 $\pm$ 0.5} & 82.0 $\pm$ 0.2 & \textbf{82.8 $\pm$ 1.0} \\
ScanObjectNN BG \cite{uy-scanobjectnn-iccv19} & 75.1 $\pm$ 0.3 & 75.6 $\pm$ 0.4 & \textbf{75.7 $\pm$ 0.5}  & 81.0 $\pm$ 0.2 & 81.8 $\pm$ 0.9 & \textbf{82.6 $\pm$ 0.7} \\
\bottomrule

\end{tabular}
}
\end{center}
\vskip -0.1in
\caption{Effect of different rendering techniques to our pre-training}\label{tab:image}
\end{table}

\section{Details of downstream tasks}
\subsection{Object classification}
\label{sec:cls}
Similar to the comparison with the DGCNN backbone in the main paper, we provide comparisons with the PointNet backbone. The results are shown in Table \ref{tab:classification}. As can be seen, our method outperforms random inititalization as well as other pre-training methods, including Jigsaw~\cite{sauder2019self}, OcCo~\cite{wang2021unsupervised}, and CM~\cite{jing2021self}.
\begin{table}[t!]
\begin{center}
\resizebox{0.8\linewidth}{!}{
\begin{tabular}{l ccccc}
\toprule
    \multirow{2}{*}{} &
      \multicolumn{5}{c}{PointNet}\\
        \cmidrule{2-6}
    &  Random & Jigsaw & OcCo & CM & Ours \\
\midrule
MN40 \cite{Wu_2015_CVPR} & 88.9{\scriptsize \textcolor{black}{$\pm$0.0}} & 89.2{\scriptsize \textcolor{black}{$\pm$0.0}} & 89.2{\scriptsize \textcolor{black}{$\pm$0.1}}& 89.1{\scriptsize \textcolor{black}{$\pm$0.1}} & \textbf{89.5{\scriptsize \textcolor{black}{$\pm$0.2}}} \\
SO \cite{uy-scanobjectnn-iccv19} & 78.2{\scriptsize \textcolor{black}{$\pm$0.1}}& 79.4{\scriptsize \textcolor{black}{$\pm$0.1}} & 79.5{\scriptsize \textcolor{black}{$\pm$0.1}} & 79.3{\scriptsize \textcolor{black}{$\pm$0.5}} & \textbf{80.5{\scriptsize \textcolor{black}{$\pm$0.4}}}   \\
SO BG \cite{uy-scanobjectnn-iccv19} & 76.4{\scriptsize \textcolor{black}{$\pm$0.0}} & 76.4{\scriptsize \textcolor{black}{$\pm$0.4}}& 76.4{\scriptsize \textcolor{black}{$\pm$0.1}} & 74.1{\scriptsize \textcolor{black}{$\pm$0.2}} &  \textbf{78.5{\scriptsize \textcolor{black}{$\pm$0.5}}} \\
\bottomrule
\end{tabular}
}
\end{center}
\vskip -0.1in
\caption{Comparison among random, Jigsaw~\cite{sauder2019self}, OcCo~\cite{wang2021unsupervised}, CM~\cite{jing2021self}, and our initialization to the object classification downstream task. We reported the mean and standard deviation at the best epoch over three runs.}\label{tab:classification}
\vskip -0.1in
\end{table}

\subsection{A note on the OcCo baseline} 
It can be seen that in our paper, some experiment results of OcCo are lower than the results reported by its original paper.
We did our best to reproduce the results of OcCo but unfortunately, we were not able to match the results with the original paper. 
We confirmed this issue by using the docker image provided by the OcCo authors and rerun the experiments, but still could not reproduce the results exactly as in the OcCo paper. 
For fair comparison and reproducibility, we decided to report the results based on our own runs.  
Additionally, the pre-training time of OcCo is about 7x slower than our method.

\subsection{Semantic segmentation}
\label{sec:semseg} 

In additional to the Area-5 results reported in the main paper, we further report the results of 6-fold cross-validation over the 6 areas on the S3DIS dataset. 
For completeness, all results are shown in Table \ref{tab:seemseg_area5} (Area-5), and Table~\ref{tab:seemseg} (6 folds). 
In both cases, we can see that models initialized by our method outperform others in both PointNet~\cite{Qi_2017_CVPR} and DGCNN~\cite{wang2019dynamic}.

\begin{table*}[t!]
\setlength\tabcolsep{6pt}
\begin{center}
\begin{tabular}{c cccc|cccc}
\toprule
    \multirow{2}{*}{} &
      \multicolumn{4}{c|}{PointNet} &
      \multicolumn{4}{c}{DGCNN}\\
        \cmidrule{2-9}
    & Random & Jigsaw & OcCo & Ours & Random & Jigsaw & OcCo & Ours \\
\midrule
mAcc & 83.9 & 82.5 & 83.6 & \textbf{85.0} & 86.8 &  86.8 & \textbf{87.0} & \textbf{87.0} \\
mIoU & 43.6 & 43.6 & 44.5 & \textbf{46.7} & 49.3 & 48.2 & 49.5 & \textbf{49.9} \\
\bottomrule
\end{tabular}
\end{center}
\vskip -0.1in
\caption{Fold 1 of overall point accuracy (mAcc) and mean Intersection-over-Union (mIoU) on the S3DIS (Stanford Area 5 Test) \cite{armeni20163d}}
\label{tab:seemseg_area5}
\end{table*}

\begin{table*}[t!]
\begin{center}
\begin{tabular}{l cccc|cccc}
\toprule
    \multirow{2}{*}{} &
      \multicolumn{4}{c|}{PointNet} &
      \multicolumn{4}{c}{DGCNN}\\
        \cmidrule{2-9}
    & Random & Jigsaw & OcCo & Ours & Random & Jigsaw & OcCo & Ours \\
\midrule
mAcc & 82.8 & 82.8 & 82.7 & \textbf{83.2} & 86.9 &  86.6 & 87.1 & \textbf{87.5} \\
mIoU & 50.6 & 51.4 & 51.1 & \textbf{52.1} & 58.4 & 58.1 & 58.7 & \textbf{59.0} \\
\bottomrule
\end{tabular}
\end{center}
\vskip -0.1in
\caption{Average of 6-fold cross validation of overall point accuracy (mAcc) and mean Intersection-over-Union (mIoU) on the S3DIS \cite{armeni20163d}}
\label{tab:seemseg}
\end{table*}

\subsection{Details of PointContrast baseline}
\label{sec:pointcontrast}
\noindent\textbf{Semantic segmentation: } We evaluate on two datasets S3DIS \cite{armeni20163d} and ScanNet \cite{dai2017scannet}. We use SGD optimizer with the initial learning rate 0.1 and 0.8 for S3DIS and ScanNet respectively. We use PolynomialLR scheduler with a power factor 0.9.For ScanNet dataset, we train the model with 15000 iterations and batch size 48 on 4 GPUs. For S3DIS dataset, we train the model with 20000 iterations and batch size 32 on 1 GPU.



\noindent\textbf{Object detection: } For object detection task, in the training  we follow the configuration of PointContrast~\cite{xie2020pointcontrast}. We use Adam optimizer with the initial learning rate 0.001 and train the model on 1 GPU with 180 epochs. Specifically, we train the model with batch size 32 and 64 for ScanNet and SUN RGB-D, respectively. Before voxelization, we sample 40000 and 20000 points from original point of ScanNet and SUN RGB-D and the voxel sizes are 2.5 cm and 5 cm respectively. 

\begin{table*}[h]
\begin{center}
\begin{tabular}{l cccc|cccc}
\toprule
    \multirow{2}{*}{ModelNet40~\cite{Wu_2015_CVPR}} &
      \multicolumn{4}{c|}{PointNet} &
      \multicolumn{4}{c}{DGCNN}\\
        \cmidrule{2-9}
    & Random & Jigsaw & OcCo & Ours & Random & Jigsaw & OcCo & Ours \\
\midrule
5\% & 73.2 & 73.8 & 73.9 & \textbf{77.9} & 82.0 & 82.1 & 82.3 & \textbf{84.9} \\
10\% & 75.2 & 77.3 & 75.6 & \textbf{79.0} & 84.7 & 84.1 & 84.9 & \textbf{86.6} \\
20\% & 81.3 & 82.9 & 81.6 & \textbf{84.6} & 89.4 & 89.2 & 89.1 & \textbf{90.2} \\
50\% & 86.6 & 86.5 & 87.1 & \textbf{87.6} & 91.6 & 91.8 & 91.7 & \textbf{92.4} \\
70 \% & 88.3 & 88.4 & 88.4 & \textbf{88.7} & 92.3 & 92.4 & 92.5 & \textbf{92.8} \\
\midrule
    \multirow{2}{*}{ScanObjectNN \cite{uy-scanobjectnn-iccv19}} &
      \multicolumn{4}{c|}{PointNet} &
      \multicolumn{4}{c}{DGCNN}\\
        \cmidrule{2-9}
    & Random & Jigsaw & OcCo & Ours & Random & Jigsaw & OcCo & Ours \\
\midrule
5\% & 52.1 & 51.8 & 53.7 & \textbf{60.8} & 48.3 & 46.7 & 51.4 & \textbf{60.9} \\
10\% & 63.0 & 62.3 & 62.5 & \textbf{69.0} & 58.7 & 58.0 & 61.5 & \textbf{69.5} \\
20\% & 69.0 & 68.5 & 67.1 & \textbf{72.2} & 69.8 & 68.7 & 71.6 & \textbf{74.7} \\
50\% & 73.7 & 75.1 & 72.6 & \textbf{77.0} & 76.3 & 77.1 & 78.0 & \textbf{81.6} \\
80 \% & 76.1 & 77.9 & 76.7 & \textbf{78.4} & 79.9 & 78.1 & 80.8 & \textbf{82.1} \\
\bottomrule
\end{tabular}
\end{center}
\vskip -0.1in
\caption{Performance of the object classification task trained with fewer data on ModelNet40 \cite{Wu_2015_CVPR} and ScanObjectNN \cite{uy-scanobjectnn-iccv19}. Our method has significant gains compared to other initialization methods. We reported the mean at the best epoch over three runs}\label{tab:ratio}
\end{table*}


\begin{figure}
\resizebox{\linewidth}{!}{
\begin{subfigure}{.33\textwidth}
  \centering
  \includegraphics[width=1\linewidth]{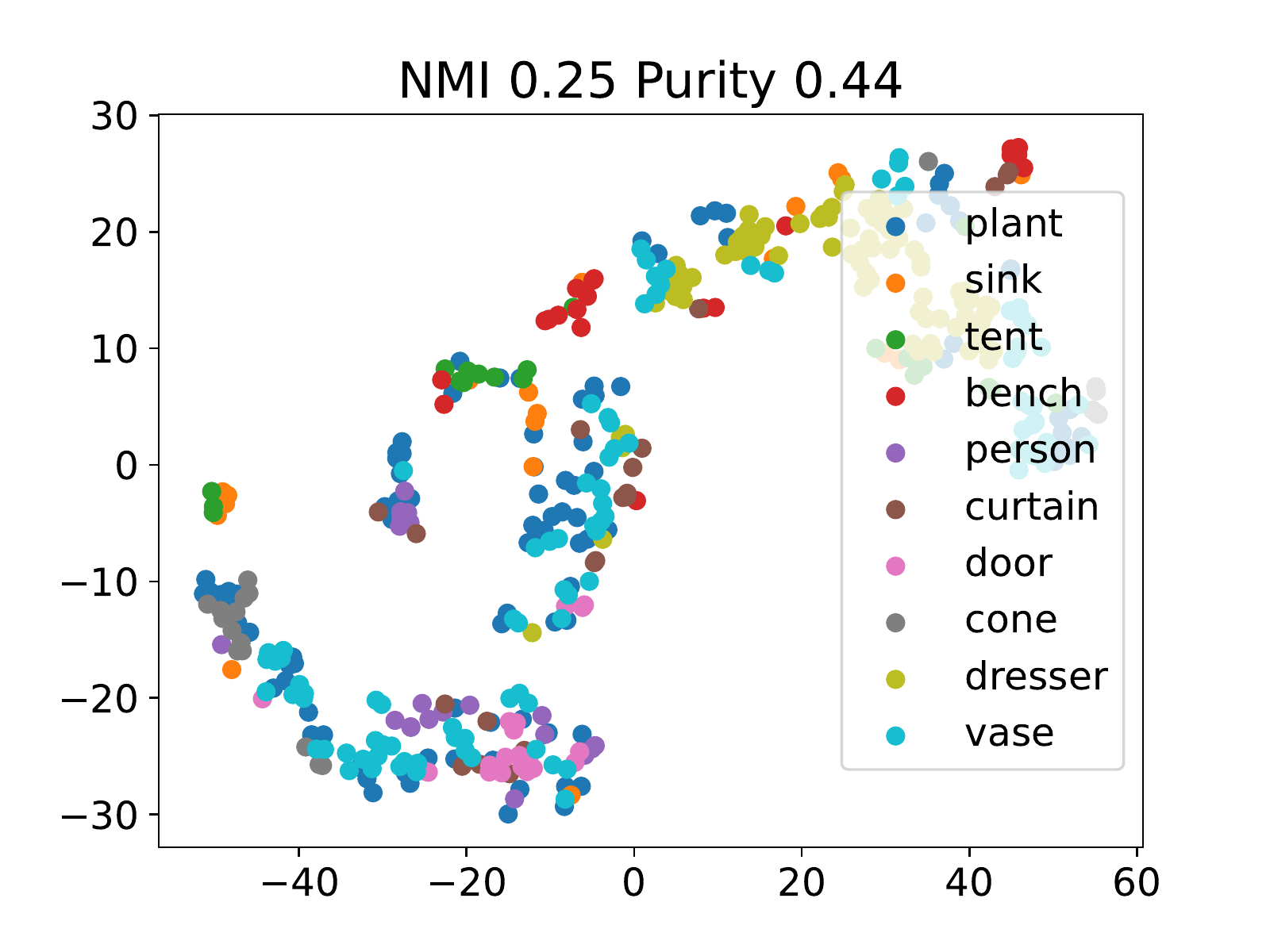}  
  \caption{Jigsaw~\cite{sauder2019self} on PointNet}
\end{subfigure}
\begin{subfigure}{.33\textwidth}
  \centering
  \includegraphics[width=1\linewidth]{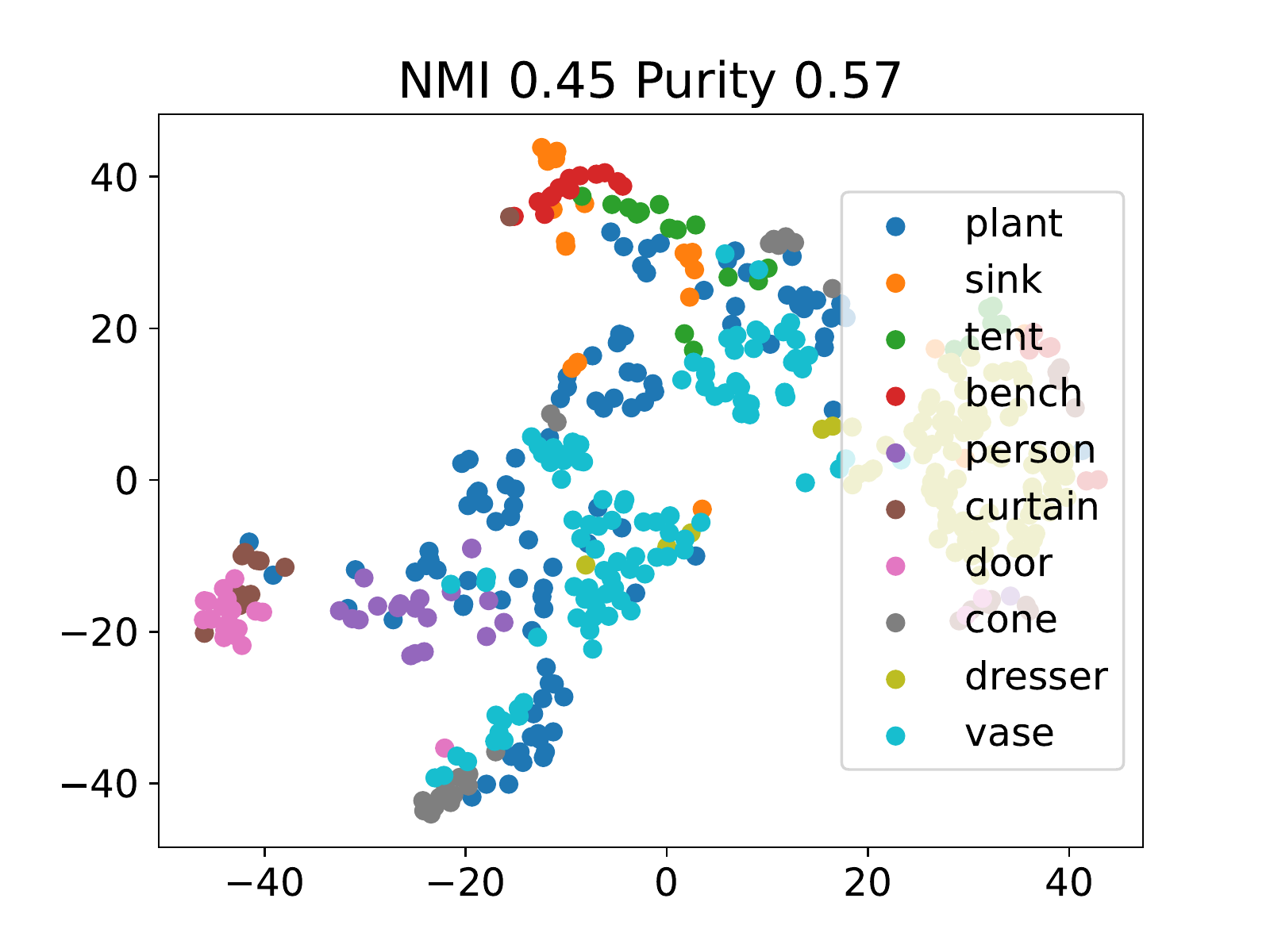}  
  \caption{OcCo~\cite{wang2021unsupervised} on PointNet}
\end{subfigure}
\begin{subfigure}{.33\textwidth}
  \centering
  \includegraphics[width=1\linewidth]{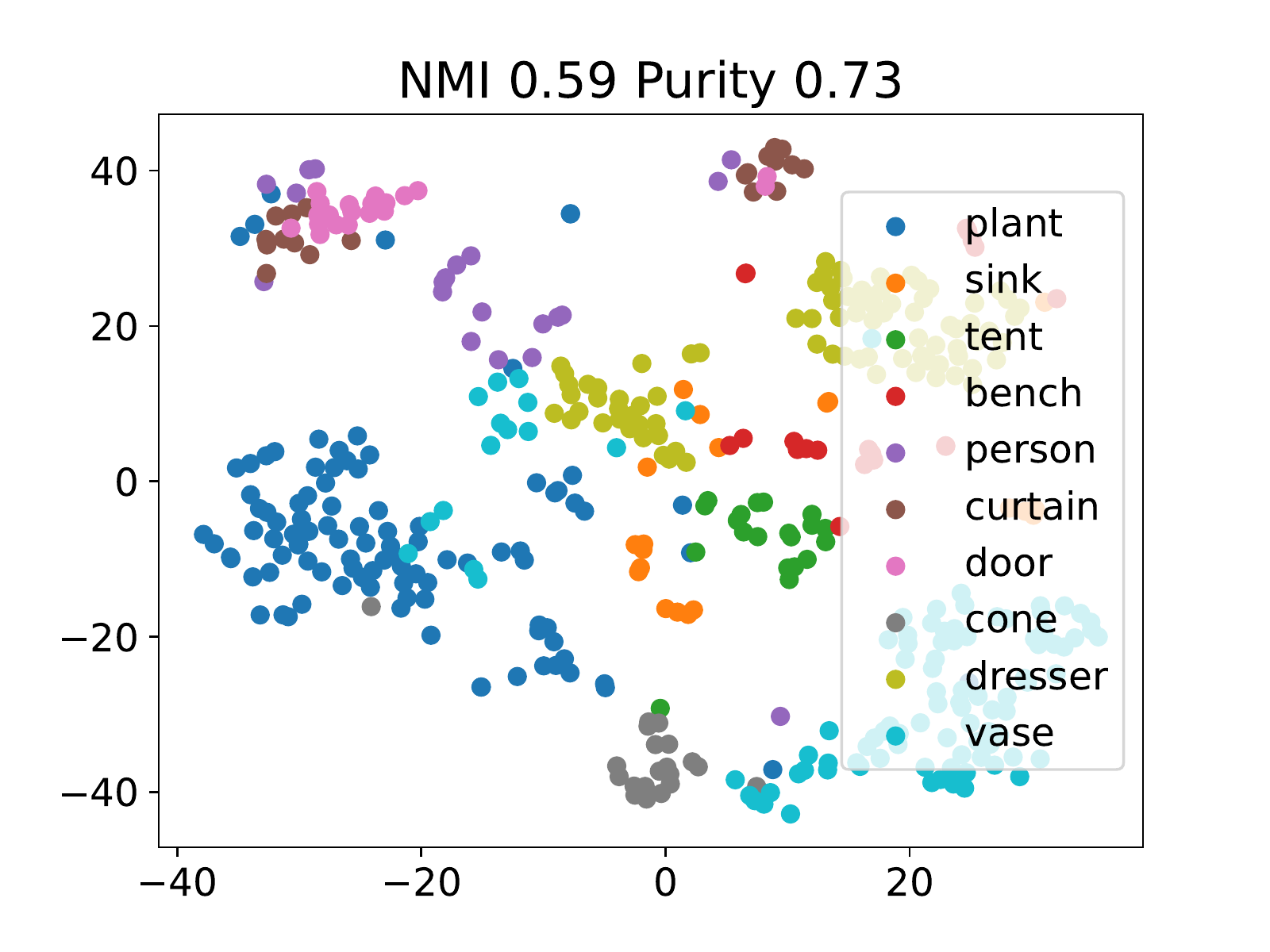}  
  \caption{Ours on PointNet}
\end{subfigure}
}
\newline
\resizebox{\linewidth}{!}{
\begin{subfigure}{.33\textwidth}
  \centering
  \includegraphics[width=1\linewidth]{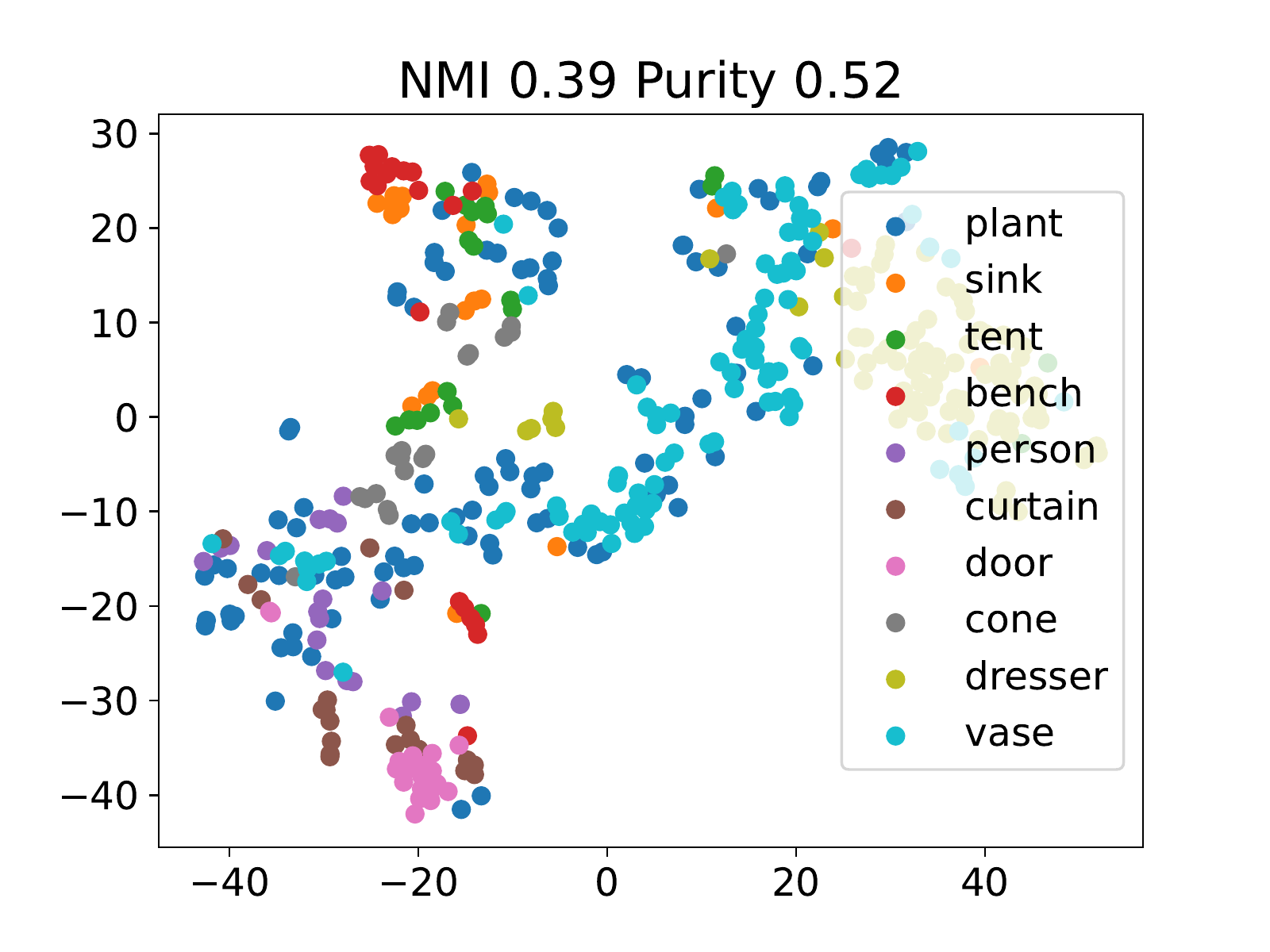}  
  \caption{Jigsaw~\cite{sauder2019self} on DGCNN}
\end{subfigure}
\begin{subfigure}{.33\textwidth}
  \centering
  \includegraphics[width=1\linewidth]{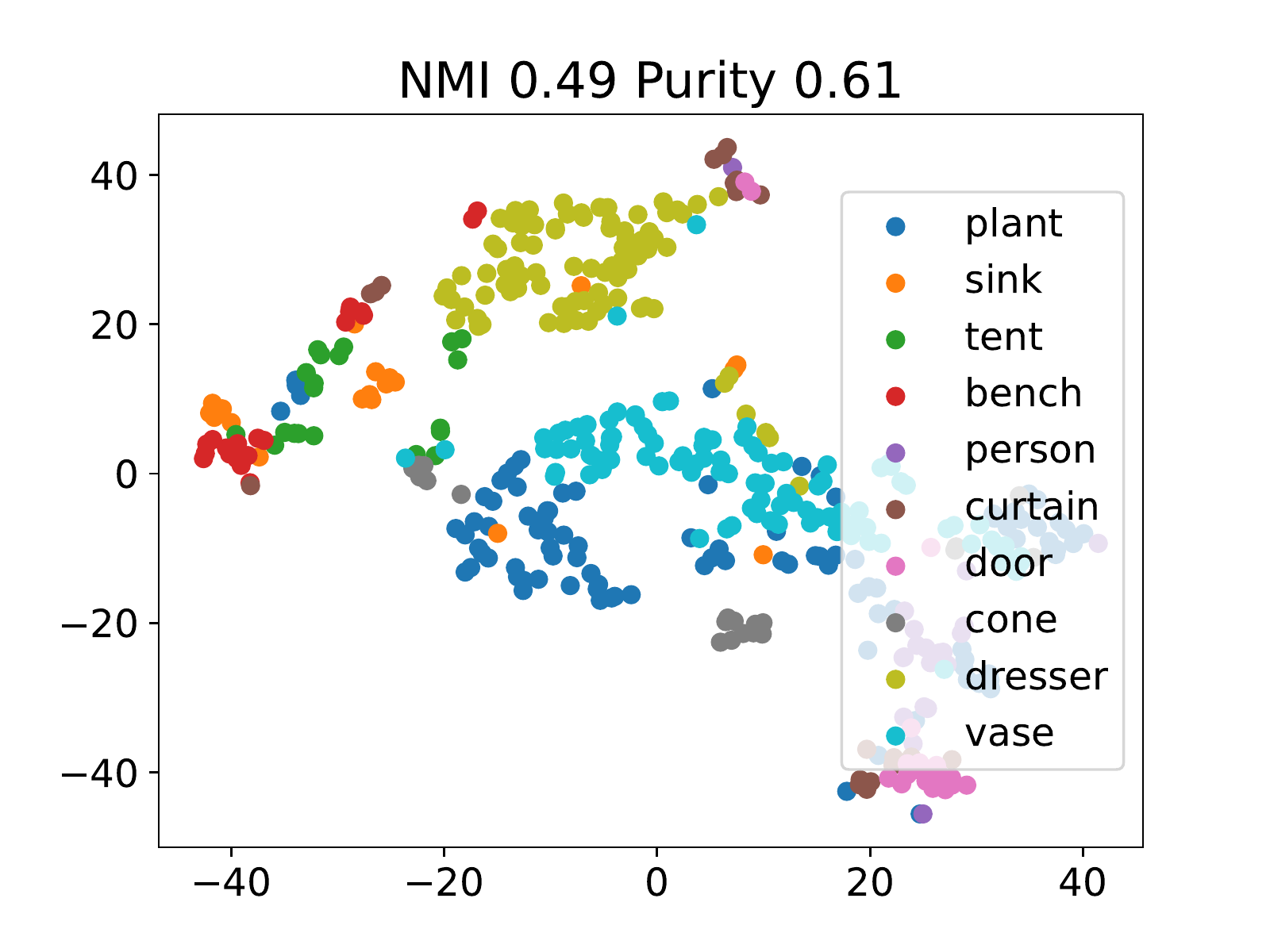}  
  \caption{OcCo~\cite{wang2021unsupervised} on DGCNN}
\end{subfigure}
\begin{subfigure}{.33\textwidth}
  \centering
  \includegraphics[width=1\linewidth]{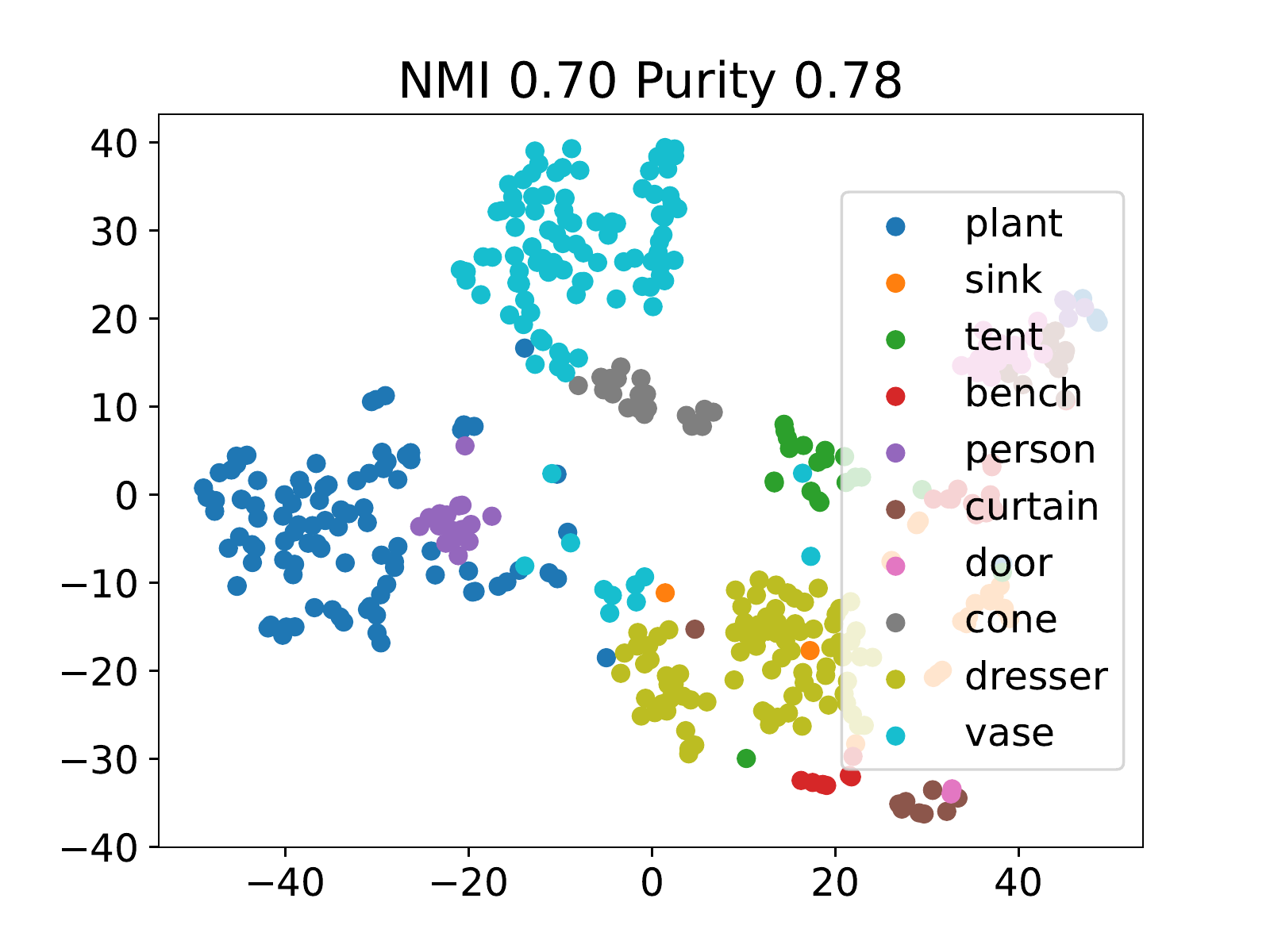}  
  \caption{Ours on DGCNN}
\end{subfigure}
}
\vskip -0.1in
\caption{t-SNE visualization of the object embedding of the test data of ModelNet40. Our method has better cluster quality measured by NMI and purity.}
\label{fig:visual_feature}

\end{figure}

\begin{figure}
\begin{center}
\includegraphics[width=\textwidth]{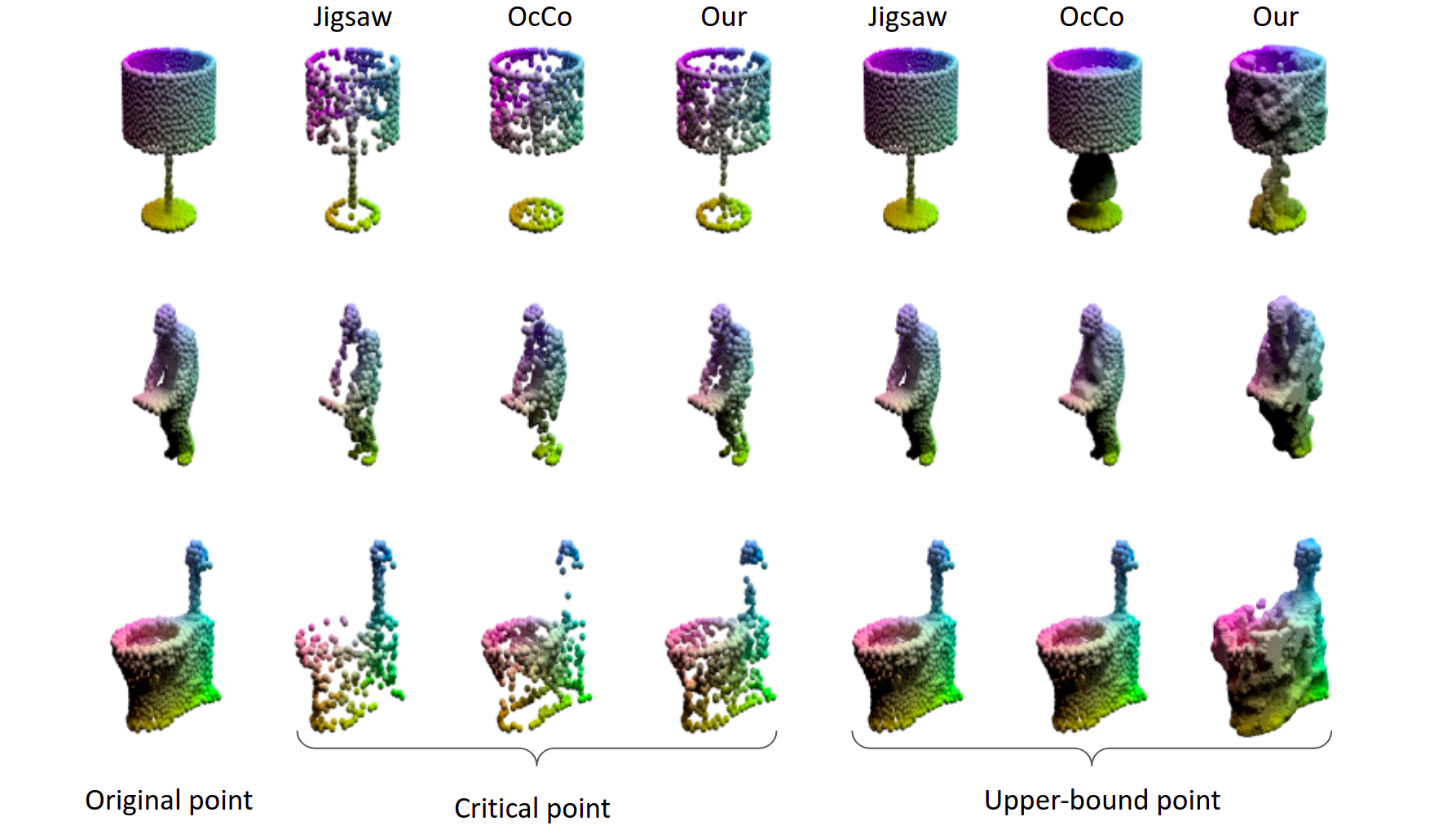}
\caption{Critical and upper-bound point visualizations for models pretrained with Jigsaw, OcCo, and our method.}
\label{fig:critical_point}
\end{center}
\end{figure}
\section{Training with limited data} 
\label{sec:limited}

To prove the effectiveness of our pre-training, we supervise the downstream task with fewer data when the network is pre-trained and compare to other initializations. 
We show both results on ModelNet40 (also reported in the paper) and ScanObjectNN. 
In this experiment, we decrease the number of training samples to 5\%, 10\%, 20\%, 50\%, and 80\%, and evaluate the model on the original test set. 
The results are reported in Table~\ref{tab:ratio}, which shows that the performance of our method outperforms Random, Jigsaw \cite{sauder2019self}, and OcCo \cite{wang2021unsupervised} in most cases except DGCNN on 80\% of ScanObjectNN.

\section{Visualization}
\label{sec:visual} 

\subsection{t-SNE embedding}
We further visualize learned object embeddings of the ModelNet40 test set by using t-SNE with perplexity 15 and 1000 iterations in Figure \ref{fig:visual_feature}. We observe that the embeddings learned from using our initialization for different classes are well clustered than those acquired with OcCo and Jigsaw initialization indicated by normalized mutual information (NMI) and purity~\cite{manning2008introduction}. 

\subsection{Critical point sets}
We then visualize the critical point sets and upper-bound shapes by following PointNet~\cite{Qi_2017_CVPR} for selected samples in Figure \ref{fig:critical_point}. 
To recap, a critical point set is a set of points that contribute directly to the last max pooled feature, i.e., the global feature. Perturbing the critical point set can lead to changes in the global features and thus classification results. 
The upper-bound shape is the largest possible point set that does not directly affect the global feature of the original shape. 
From Figure \ref{fig:critical_point}, we found that in our method, the critical point sets can represent the object skeleton more faithfully (e.g., the toilet example) than other methods. 
Jigsaw sometimes causes sparse critical points, and OcCo tends to discard points along thin geometric features.
We also found that the upper-bound shape of our initialization appears thicker than that of Jigsaw and OcCo, which we hypothesize that our model can be more robust to point perturbations than Jigsaw and OcCo.


\section{Running time} 
Following the request, we provide the pre-training time of three methods on \textbf{an NVIDIA Tesla V100 GPU} in Table~\ref{tab:pretrain_time}. As can be seen, the pre-training time of our method is slightly longer than Jigsaw and significantly shorter than OcCo. Despite such, our method achieves better performance than the others.

\begin{table*}[h]
\setlength\tabcolsep{5pt}
\begin{center}
\resizebox{0.4\linewidth}{!}{
\begin{tabular}{l cccc}
\toprule
    & Jigsaw & OcCo & Ours \\
\midrule
PointNet & 2.6 & 24.8 & 3.8 \\
DGCNN & 4.1 & 35.1 & 5.7 \\
\bottomrule
\end{tabular}
}
\end{center}
\vskip -0.1in
\caption{Pre-training time of three methods (hours).}\label{tab:time}
\label{tab:pretrain_time}
\vskip -0.4in
\end{table*}

Additionally, we provide more statistics of our training process. 
Specifically, it takes 2.3, 2.4, and 6.2 hours to render RGB, silhouette, and shaded images, respectively. 
For the 2D self-supervision, we train the model for 80 hours on an NVIDIA Tesla V100 GPU. 
Knowledge distillation takes 3.8, 5.7, 26 and 62  hours of training for PointNet \cite{Qi_2017_CVPR}, DGCNN \cite{wang2019dynamic}, SR-UNet on ModelNet40~\cite{Wu_2015_CVPR} and SR-UNet on ScanNet~\cite{dai2017scannet}, respectively.
As for downstream-task training, the PointNet classification model takes 18.5 hours, and the DGCNN classification model takes 75.0 hours. The segmentation models require longer training time, with 32.0 hours and 90.0 hours for PointNet and DGCNN backbone, respectively. 
For SR-UNet backbone\cite{choy20194d}, in semantic segmentation task, it consumes 32 and 22 hours for S3DIS~\cite{armeni20163d} and ScanNet~\cite{dai2017scannet}, respectively. For object detection task on ScanNet~\cite{dai2017scannet}, it takes 8.5 hours.

\section{Future Work}
Our method is not without limitations. 
First, our image encoder is trained from scratch without leveraging existing popular feature extractors such as VGG~\cite{simonyan2014very} or ResNet~\cite{he2016deep}. 
Further utilizing such pre-trained networks on natural images could potentially improve the performance of the downstream tasks, which could be interesting for future work. 
Second, the multi-view rendering used in our method could potentially be further explored. While we attempted with position encoding, silhouette, and shaded rendering, there are many other rendering styles that could be experimented, e.g., rendering with colors and textures when applicable, rendering with depth completion, etc. Applying advanced techniques to enhance multi-view rendering is thus a good avenue for future research.

\end{document}